\documentclass{article} 
\usepackage{iclr2025_conference,times}

\usepackage{amsmath,amsfonts,bm}

\def\eqref#1{equation~\ref{#1}}

\def\1{\bm{1}}

\DeclareMathAlphabet{\mathsfit}{\encodingdefault}{\sfdefault}{m}{sl}
\SetMathAlphabet{\mathsfit}{bold}{\encodingdefault}{\sfdefault}{bx}{n}

\usepackage{hyperref}
\usepackage{url}
\usepackage{tabularx, colortbl,booktabs}
\usepackage{xspace}
\usepackage[pdftex]{graphicx}
\usepackage{enumitem}
\usepackage{listings}
\usepackage{pifont}
\title{Planning Anything with Rigor: General-Purpose Zero-Shot Planning with LLM-based Formalized Programming}

\author{Yilun Hao \\
  MIT \\
  \texttt{yilunhao@mit.edu} \\\And
  Yang Zhang \\
  MIT-IBM Watson AI Lab \\
  \texttt{Yang.Zhang2@ibm.com}\\\And
  Chuchu Fan \\
  MIT\\
  \texttt{chuchu@mit.edu}\\}

\newcommand{\ours}{LLMFP\xspace}
\newcommand{\oursfull}{\textbf{LLM}-Based \textbf{F}ormalized \textbf{P}rogramming\xspace}
\newcommand{\rom}[1]{\uppercase\expandafter{\romannumeral #1\relax}}

\newcommand{\blue}[1]{\textcolor{black}{#1}}

\iclrfinalcopy 
\begin{document}

\maketitle
\begin{abstract}
While large language models (LLMs) have recently demonstrated strong potential in solving planning problems, there is a trade-off between flexibility and complexity. LLMs, as zero-shot planners themselves, are still not capable of directly generating valid plans for complex planning problems such as multi-constraint or long-horizon tasks. On the other hand, many frameworks aiming to solve complex planning problems often rely on task-specific preparatory efforts, such as task-specific in-context examples and pre-defined critics/verifiers, which limits their cross-task generalization capability. In this paper, we tackle these challenges by observing that the core of many planning problems lies in optimization problems: searching for the optimal solution (best plan) with goals subject to constraints (preconditions and effects of decisions). With LLMs' commonsense, reasoning, and programming capabilities, this opens up the possibilities of a universal LLM-based approach to planning problems. Inspired by this observation, we propose \ours, a general-purpose framework that leverages LLMs to capture key information from planning problems and formally formulate and solve them as optimization problems from scratch, with no task-specific examples needed. We apply \ours to 9 planning problems, ranging from multi-constraint decision making to multi-step planning problems, and demonstrate that \ours achieves on average 83.7\% and 86.8\% optimal rate across 9 tasks for GPT-4o and Claude 3.5 Sonnet, significantly outperforming the best baseline (direct planning with OpenAI o1-preview) with 37.6\% and 40.7\% improvements. We also validate components of \ours with ablation experiments and analyzed the underlying success and failure reasons. Project page: \href{https://sites.google.com/view/llmfp}{https://sites.google.com/view/llmfp}.
\end{abstract}

\section{Introduction}
\label{sec:intro}
\begin{figure}[t]
  \includegraphics[width=\linewidth]{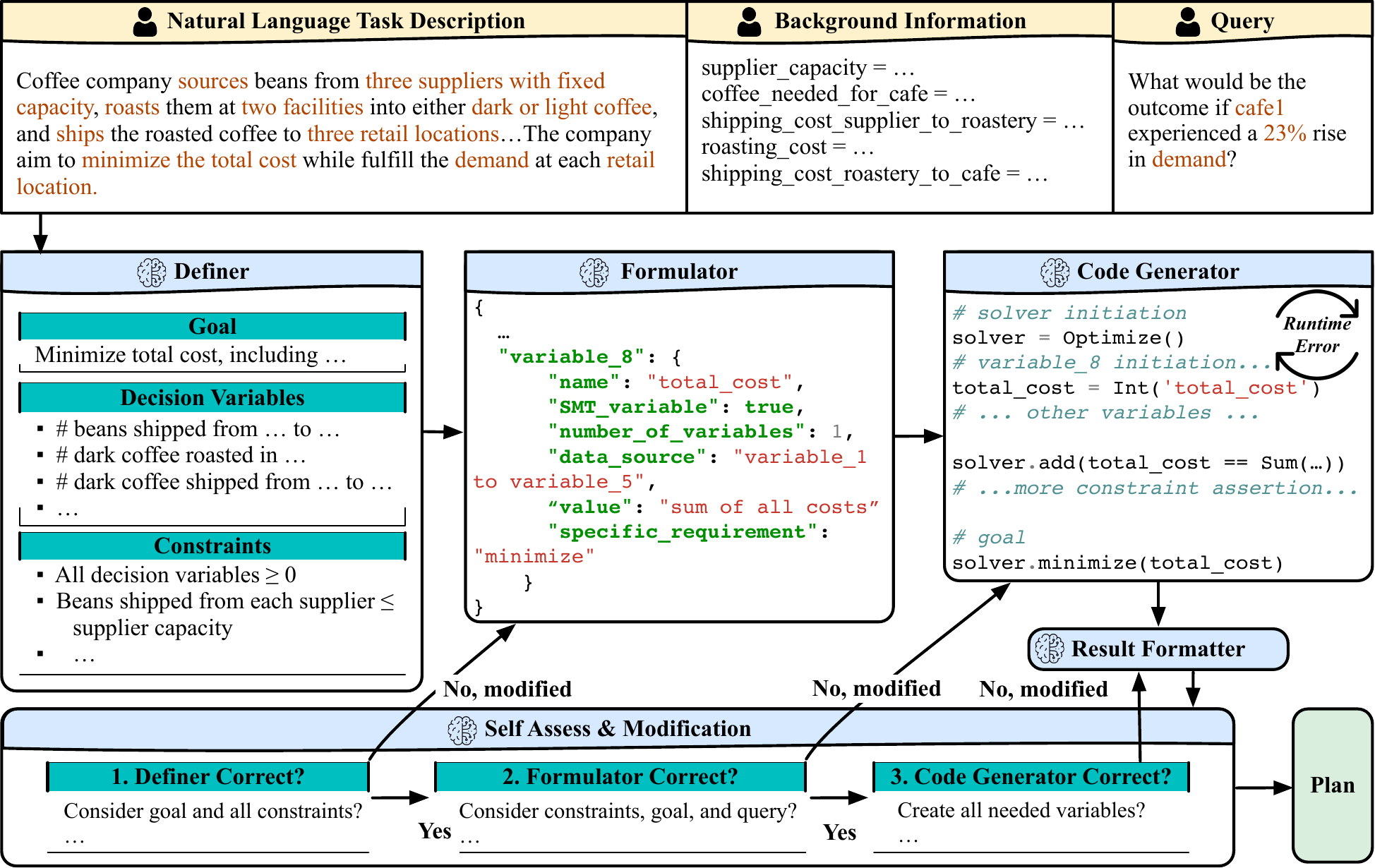} 
  \caption {An overview of \ours and how it is applied to a coffee supply chain example. All sections in yellow are inputs, and all sections in blue are steps accomplished by LLMs. With task description, background information, and query as inputs, \ours defines the goal, decision variables, and constraints of this optimization problem, identifies all necessary variables and summarizes their key information into a JSON representation, generates codes to solve the optimization problem, executes the codes and formats the execution results, and performs self-assessment for each step.}
  \label{fig:Framework}
\end{figure}


Making complex plans subject to multiple constraints is a time- and labor-intensive process, but is critical in many aspects of our lives such as work arrangement, business management, logistics, and robotics. In the past, people used domain-specific tools and languages to make specific plans in their areas, which often required a steep learning curve and were hard to adapt to other domains. When large language models (LLMs) emerge with their versatile capabilities such as language understanding, reasoning, and tool-using, using LLMs for planning has gained significant traction.

For such planning systems to be deployed in complex, real-world applications, two desirable properties need to be satisfied: 
1). \emph{Zero-shot flexibility}:
Unlike many experimental settings where planning tasks usually come with labeled datasets, it is very challenging to request such datasets from users in many realistic settings. Ideally, a flexible planning system should be able to conduct planning with only a task description provided by users, and nothing else. 
2). \emph{High performance on complex tasks}:
Realistic planning problems usually require multi-step, long-horizon solutions, with many explicit and implicit constraints.

However, there is a trade-off between flexibility and task complexity, and thus existing LLM-based planning systems are typically unable to achieve both simultaneously.
On one hand, planning systems capable of performing zero-shot planning, utilizing the abundant knowledge and generalization capabilities in LLMs, in many successful applications can only solve single-objective tasks such as household chores, with step-by-step interactive planning and grounding \citep{huang2022language, ahn2022can, huang2022inner}.
For complex, multi-constraint, and long-horizon tasks that involve iterative trials and errors even for humans, LLMs still do not have the capabilities to generate valid plans by themselves \citep{kambhampati2024llms}. On the other hand, recent research efforts to empower LLM-based planners to solve complex tasks are often based on well-designed task-specific in-context examples and extensive task-specific pre-defined efforts \citep{liu2023llm+, xie2023translating, li2023large, song2023llm, gundawar2024robust}, impacting their zero-shot flexibility. In short, few existing planning systems can flexibly resolve generic complex tasks with only task descriptions in natural language. Hence we ask: Can we build a universal LLM-based planning system that can solve complex planning problems without task-specific efforts?

In this paper, we observe that, although planning problems come with drastic variations, many of them can be recast as constrained optimization problems --- The optimization problems aim to find the optimal solution, which is equivalent to locating the best plan that satisfies the goal for planning problems; the decision's precondition and effect are equivalent to constraints of optimization problems. Furthermore, although solving complex planning tasks is generally beyond the capabilities of LLMs, converting any planning tasks into optimization problems is a much more tractable problem, and can be within the zero-shot capabilities of LLMs.

Motivated by this, we propose \oursfull (\ours, illustrated in Fig.~\ref{fig:Framework}), a general-purpose zero-shot planning framework that leverages LLM's strong common reasoning, and programming capabilities to encode planning problems into optimization problems \emph{without} any task-specific examples or designs, combined with a formal planner to solve the optimization problem.
\ours takes in natural language domain description, natural language query under this domain, and available background information or APIs as inputs, and solves the planning problem in five steps. First, \ours prompts LLMs to reason and propose the goal, decision variables, and key constraints necessary for the task. Second, based on the response, \ours asks LLMs to formulate a representation that includes all variables needed to construct and their detailed information and requirements. Third, with the representation, LLMs write codes to formally encode the problem into an optimization problem. Fourth, \ours executes the generated codes, converts the execution results into plans. Finally, \ours performs overall self-assessment and automatic modification to fix the broken parts of the previous steps. Currently, \ours uses the satisfiability modulo theory (SMT) to encode the optimization problems but can be adapted to any planners or solvers by updating the requirements and representation format in the prompts.


We evaluate our framework with 9 diverse planning problems, ranging from single-step supply chain problem to multi-step robot block stacking and moving. Experiment results demonstrate that \ours achieves strong performance across all tasks with an average of 83.7\% and 86.8\% optimal rates for GPT-4o and Claude 3.5 Sonnet, which greatly outperforms the baselines, including direct plan generation with OpenAI o1-preview. We conduct ablation experiments to validate the key components of \ours and investigate the underlying reasons why it is more effective than baselines. 
In addition, although our framework does not require task-specific examples, we show the ease of adding task-specific examples to one stage of \ours, and how it could help to clarify unclear queries and therefore can further improve the performance within the same domain.

In summary, our key contributions are:
\begin{itemize}[left = 0pt, noitemsep,topsep=0pt,parsep=2pt,partopsep=2pt] 
\setlength\itemsep{0em}
  \item We offer a novel perspective on using LLMs to solve planning problems by rigorously constructing optimization problems from scratch, alike how human experts use optimization tools for planning.
  \item We propose \ours, a general-purpose planning framework with zero-shot generalization capability. To our knowledge, \ours is the first to enable LLMs to build and solve diverse types of planning problems as optimization problems with no task-specific examples or external critics.
  \item \ours notably achieves 83.7\% and 86.8\% optimal rates for GPT-4o and Claude 3.5 Sonnet, outperforming the best baseline (direct planning with OpenAI o1-preview) by 37.6\% and 40.7\%. We examine the effectiveness of our framework and analyze the success and failure reasons.
\end{itemize}
\vspace{-\topsep}

\section{Related Works}
\label{sec:related_work}
\subsection{LLMs for Planning}
The remarkable capabilities of LLMs in reasoning\blue{~\citep{wei2022chain, kojima2022large, yao2022react, yao2024tree, raman2024cape}} and tool-use~\citep{qin2023toolllm, schick2024toolformer} brings up interests of utilizing LLMs for planning problems. Based on LLMs's zero-shot generalization capability, many works are proposed to perform zero-shot planning~\citep{huang2022language, ahn2022can, huang2022inner}. However, their planning scenarios are limited to single-objective tasks such as household cleaning and often require step-by-step interactive planning with grounding. To improve LLMs planning capabilities for complex problems, chain-of-thought (CoT) prompting LLMs to perform step-by-step reasoning~\citep{wei2022chain}. Recent works also propose to aid the LLM planning with external tools~\citep{liu2023llm+, guan2023leveraging, chen2023autotamp, li2023large, gundawar2024robust, hao2024large, chen2024prompt}. For example, \citep{liu2023llm+, xie2023translating, gundawar2024robust} leverages LLMs to translates problems into fixed formats and solve with external planners, \citep{li2023large} prompts LLM to add short codes to existing codes to account for follow-up what-if questions, and \blue{\citep{gundawar2024robust,chen2024prompt}} empowers LLMs to iteratively refine plans or prompts based on feedback from external task-specific critics/verifiers/humans. However, to achieve strong performance, these methods often need extensive task-specific pre-defined efforts. For example, CoT depends on task-specific examples to achieve strong performance, planning domain definition language (PDDL) domain files~\citep{aeronautiques1998pddl, haslum2019introduction} are required for \citep{liu2023llm+}, mixed-integer linear program (MILP) codes of current domains are necessary for \citep{li2023large}, and external constraint critics are needed for \citep{gundawar2024robust}. These requirements limit their generalization capabilities to new domains.

\subsection{LLM + Solver}
As existing LLMs do not have the capability to perform long-horizon reasoning for complex tasks~\citep{achiam2023gpt, valmeekam2022large, valmeekam2023planning, kambhampati2024llms}, many works propose to take advantages of both LLMs and external solvers by combining them for reasoning or planning. \citep{wu2022autoformalization, he2023solving, pan2023logic, ye2024satlm} combines LLM with symbolic solvers to solve logical reasoning problems. While most logical reasoning problems are single-step satisfaction problems with clear constraint descriptions, \ours aims to solve complex planning problems, which could include implicit constraints not described in the task description and could be sequentially long-horizon tasks with defined actions. In addition, \ours proposes a general approach, which does not require task-specific examples or task-specific efforts. \citep{li2023large} teaches LLMs to add constraints to existing MILP codes. \citep{li2024formal} asks the developer to express planning problems into automaton and plan based on it. \blue{\citep{manas2024cot} uses LLMs to translate problem into linear temporal logic and solves with optimization solver.} \citep{liu2023llm+,guan2023leveraging, zhou2024isr, xie2023translating} leverages PDDL planner to aid the planning processes. Except for the natural language task description, they require human efforts to design solver-related specifications and task-specific examples, which is not needed for \ours. 

\section{\ours}
\label{sec:method}

\ours aims to resolve generic planning problems. For example, consider a \textit{coffee supply chain problem}, where a coffee company sources beans from three suppliers with fixed capacity, roasts them at two facilities into dark or light coffee and ships the roasted coffee to three retail locations to fulfill their demands. Then a planning problem involves accomplishing the task at the cheapest cost.

To achieve this, \ours takes the following inputs from users, as shown in Figure~\ref{fig:Framework} (top panels).
\begin{itemize}[left = 0pt, noitemsep,topsep=0pt,parsep=2pt,partopsep=2pt] 
\item \textbf{Natural Language Task Description $\boldsymbol{d}$.} A natural language description that details the problem settings and the planning objective, such as the above description of the coffee problem.
\item \textbf{Background Information \& API $\boldsymbol{i}$.} A list of background information about the tasks as well as information on APIs that the planner can use. An example of the background information for the coffee task is the variables containing specific numbers of supplier capacities, cafe demands, and costs for shipping and roasting.
\item \textbf{User Query $\boldsymbol{q}$.} The question that either describes the detailed initial and/or goal states or adds/modifies existing requirements of the tasks.  In the coffee planning task, one example query is \textit{`What would be the outcome if cafe1 experienced a 23\% rise in demand'}.
\end{itemize}
Example inputs for all 9 tasks can be found in Appendix~\ref{sec:appendix_example}. Note that \ours does not require any task-specific examples from the users. Considering the diversity of user requests, \ours needs to accommodate a large variety of domains, planning problem setups, user queries, constraints, and complexity levels, which poses a great challenge.


\subsection{Overview}
\blue{Devising solutions for the vast diversity of different complex planning problems seems prohibitive even for humans, let alone LLMs. However, despite the diversity of the planning problems, a planning problem can generally be cast as a constrained optimization problem. Formally, a constrained optimization, $\mathcal{P}=\{\bm x, f(\cdot), \bm g(\cdot), \bm h(\cdot)  \}$, is defined as $\min_{\bm x} {f(x)}$, s.t., $\bm g(\bm x) \leq 0, \bm h(\bm x) = 0$. $\bm x$ represents the decision variables; $f(\cdot)$ represents the objective function; $ \bm g(\cdot)$ represents the (multiple) inequality constraints; $\bm h(\cdot)$ represents the (multiple) equality constraints. In the coffee supply problem, $\bm x$ includes the amount of coffee beans shipped from each supplier to each roastery, and from each roastery to each cafe, $f$ describes the total cost of shipping, and $\bm g$ and $\bm h$ include the clearing conditions for each facility. A detailed description of the variables and constraints for the coffee supply problem can be found in Appendix~\ref{append:LLMFP_output}.}

\blue{Once a planning problem is formulated as the constrained optimization problem, it can be rigorously solved by solvers such as the SMT solver. Therefore, we propose an pipeline that solves the planning problem by converting them into constrained optimization problems and then solving them using the SMT solver. Our pipeline consists of} the following steps, as shown in Fig.~\ref{fig:Framework}. \ding{182} \textsc{\textbf{Definer:}} \blue{\ours first prompts an LLM to define the constrained optimization problem from the user inputs, $\mathcal{P}=\mathcal{D}(d,i)$} (Sec.~\ref{sec:method_define}). \ding{183} \textsc{\textbf{Formulator:}} \ours asks LLM to think about the necessary variables and steps to build when programming, and formulate a representation to summarize all key information of these variables (Sec.~\ref{sec:method_formulate}). \ding{184} \textsc{\textbf{Code Generator:}} Given this representation, LLMs generate codes that initialize all necessary variables, assert constraints, and add goals (Sec.~\ref{sec:method_generate}). \ding{185} \textsc{\textbf{Result Formatter:}} After \ours executes the generated codes, it prompts LLMs to convert the execution result into a fixed format and provide a short assessment of the execution results (Sec.~\ref{sec:method_format}). \ding{186} \textsc{\textbf{Self Assessment and Modification:}} \ours assesses each step based on the execution result, and modifies the first incorrect step (Sec.~\ref{sec:method_assess}). The generated plan is delivered when it passes self-assessments of all steps. Please refer to Appendix~\ref{sec:appendix_output} and~\ref{sec:appendix_prompt} for example outputs and prompts of all steps in \ours.

\subsection{Definer}
\label{sec:method_define}
The first step of building an optimization problem is to identify the goal, decision variables, and constraints of the problem \blue{from the user-supplied task description $d$ and background information $i$, \emph{i.e.,} $\mathcal{P} = \mathcal{D}(d,i)$}. \blue{The definer function $\mathcal{D}$ is accomplished by prompting the LLM to express in a natural language format (See Figure~\ref{fig:Framework} for an example), where the prompt skeleton includes \ding{182} a description of what goal, decision variables, and constraints mean and \ding{183} an instruction to output the aforementioned information, which is invariant across tasks.} The detailed prompt is listed in Appendix~\ref{sec:LLMFP_prompt}.

While generating the goals and decision variables are straightforward, generating the constraints is challenging, because certain constraints are not explicitly stated and can only be inferred by commonsense reasoning. We refer to these as the \emph{implicit constraints}.

For example, in the coffee supply chain task example, the implicit constraints include \textit{`the roasted coffee in each roastery does not exceed the beans it receives'}, \textit{`the shipped coffee from each roastery does not exceed the coffee it roasts'}, and importantly but easily overlooked, \textit{`all numbers of shipped and roasted beans and coffee need to be non-negative integer'}. 

To facilitate uncovering the implicit constraints, we include in the prompts (under the description of constraints) a three-step instruction to derive the constraints: \ding{182} Identify all decision variables in this task, \ding{183} for each pair of decision variables, consider relations (explicit, implicit, underlying assumption, unmentioned commonsense) between them to make sure all variables are connected, and \ding{184} provide a constraint reasoning first before answering. This effectively helps LLMs to better identify implicit constraints for multi-constraint planning problems. Since for multi-step planning problems the task description needs to explicitly define the preconditions and effects of each action, there will be no implicit constraint so this step is omitted.

\subsection{Formulator}
\label{sec:method_formulate}
\begin{figure}[t]
  \includegraphics[width=\linewidth]{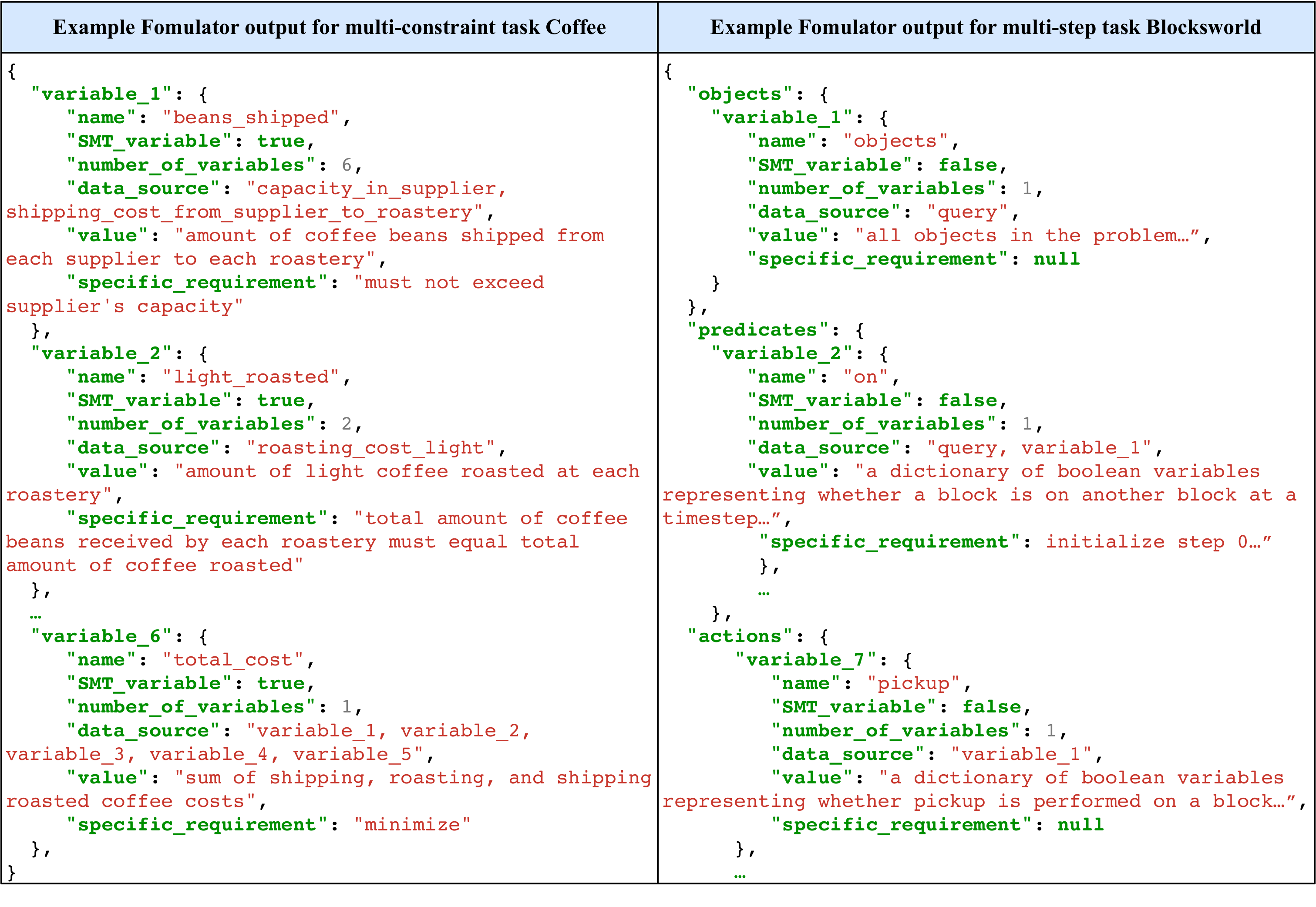} 
  \caption {Example \textsc{Formulator} output for multi-constraint Coffee and multi-step Blocksworld.}
  \label{fig:example}
\end{figure}
\blue{Before turning the optimization problem $\mathcal{P}$ into an executable code to run the SMT solver, the \textsc{Formulator} is called to supplement $\mathcal{P}$ with additional information regarding each variable in $\bm x$ that is necessary to ensure the correctness of the generated code. Formally, the \textsc{Formulator} is defined as $\mathcal{R} = \mathcal{F}(\mathcal{P}, d, i, q)$, where the output $\mathcal{R}$ is a JSON representation that describes $N$ fields of information for each variable in $\bm x$, \emph{i.e.,} $\mathcal{R}=\bigcup_i\{\bm x_i: \texttt{field}_1(\bm x_i)\}, \ldots, \texttt{field}_N(\bm x_i)\}$. Examples of the JSON representation are shown in Fig.~\ref{fig:example}.}

\blue{Each field describes the information that governs the declaration and instantiation of each variable in the code, such as whether the variable is continuous, binary, or integer, what data structures they need to be arranged into, \emph{etc.} The definition of fields is different for single-step and multi-step problems. In what follows, we will describe the fields in each problem type.}

\textbf{Single-Step Multi-Constraint Problem} \blue{As shown in Fig.~\ref{fig:Framework} and Fig.~\ref{fig:example}, for each variable, $\mathcal{R}$ includes 6 fields to summarize the information related to this variable.} The \verb|name| field indicates the variable name. Since we are using SMT as the optimization solver, the \verb|SMT_variable| field indicates whether the variable is an SMT or a normal variable. SMT variables are different from normal variables in that they don't hold specific values upfront, rather, they are symbolic variables to represent unknown values. \blue{The} \verb|number_of_variables| \blue{field represents the length of the variable. The }\verb|data_source| \blue{field denotes the dependencies of the variable. The }\verb|value| \blue{field further specifies the value of the variable. It could either be a real number or list, dictionary descriptions, or any description of operations to do with the data source. The} \verb|specific_requirement| \blue{field is where we point to the constraints or goals related to this variable. For the coffee supply chain task, the total cost has a length of 1, the} \verb|data_source| \blue{is all shipping costs and roasting costs, the} \verb|value| \blue{is a string \textit{``sum of all costs"}, and the specific requirement is \textit{``minimize"}}. With this intermediate step between \textsc{Definer} and \textsc{Code Generator}, \ours is capable of obtaining a more detailed, well-formulated, and overall coding plan. We teach LLMs to formulate variables into this representation by including two simple examples in the prompt. These two examples are not task-specific examples of any of our testing tasks and we do not modify these two examples across tasks.

\textbf{Multi-Step Planning Problem} \blue{Since multiple steps are involved in multi-step planning problems, the \textsc{Formulator} not only needs to deal with relationships between variables, but also needs to update the states of variables across different timesteps. All the variables are divided into five sections, representing the five stages for defining the variables:} \verb|objects|, \verb|predicates|, \verb|actions|, \verb|update|, \verb|goal|. \blue{The variables within each section are appended with the same six fields as introduced above.}
The \verb|objects| stage declares all objects in the scenario. The \verb|predicates| stage defines the predicates, which represent the properties of objects and the relationships between them. The \verb|actions| stage initializes variables to represent all actions. The \verb|update| stage adds assertions to existing action variables to account for the preconditions and effects of actions. The \verb|goal| stage adds constraint to existing predicate variables to encode the goal. As shown in Fig.~\ref{fig:example}, \blue{the example on the right includes different stages and variables within each stage. The variable in} \verb|predicates| \blue{stage initializes a dictionary to represent whether a block is on another block at a certain timestep. The variable in }\verb|action|\blue{ stage initializes a dictionary to represent if action pickup is performed. For multi-step planning tasks, we replace the examples with a multi-step task example, and similarly, it is not a task-specific example and we do not modify it across tasks.}

The formulator function $\mathcal{F}$ is enabled by prompting an LLM, where the prompt includes \ding{182} A brief instruction for the \textsc{Formulator}, \ding{183} Example input-output pairs of \textsc{Formulator} as demonstrations, and \ding{184} The user-provided task information and \textsc{Definer}'s output. The detailed prompt is listed in Appendix~\ref{sec:LLMFP_prompt}. Note that although example input-output pairs are used, they are task-agnostic examples fixed for all the planning tasks. No task-specific examples are needed.

\subsection{Code Generator}
\label{sec:method_generate}
With $\mathcal{R}$, now we have all the information needed to build an optimization problem with codes. In the \textsc{Code Generator}'s prompt, we explain the meanings of different stages and fields in $\mathcal{R}$ and ask LLMs to follow Python and Z3 SMT syntax~\citep{de2008z3}. By including user inputs and results from \textsc{Definer} and \textsc{Formulator}, with no examples, LLM could reliably generate reasonable, executable, and correct Python codes. Then, \ours executes the codes and returns to re-generation if there are runtime errors. We set maximum re-generation times to be 5.

\subsection{Result Formatter}
\label{sec:method_format}
Since the variable names are decided by LLMs and have chances to be very different across queries, after code generation, we use a \textsc{Result Formatter} to ask LLM to convert the execution result to a fixed output format. For example, the output for the coffee task would be a JSON includes: \ding{182} the number of coffee beans shipped from each supplier to each roastery, \ding{183} the number of light and dark coffee roasted in each roastery, and \ding{184} the number of light and dark coffee shipped from each roastery to each cafe. After filling in this result, we prompt the LLM to provide a brief evaluation of the result based on whether the result achieves the goal, satisfies constraints, and makes sense in common sense. Taking commonsense into consideration is important because sometimes if a necessary constraint is missing from the \textsc{Definer} step, it could result in unreasonable execution result that is unrealistic in commonsense. For example, for the coffee task, if the \textsc{Definer} does not include the non-negative constraint, to minimize the cost, the solver could propose negative units of shipped coffee. Detecting these unrealistic plans is helpful for \textsc{Self Assess \& Modification}. 

\subsection{Self Assess \& Modification}
\label{sec:method_assess}
With the execution result and evaluation, \ours perform self-assessment to reason about the correctness and provide ratings for the \textsc{Definer}, \textsc{Formulator}, and \textsc{Code Generator}. If the assessment marks all three steps to be correct, this plan will be delivered as the final output. Otherwise, the assessment will reason about how to modify this step, and provide a modification by itself. This modification will replace the output of the incorrect step and \ours will loop back to continue the next steps from there again. That is, if the \textsc{self Assessor} thinks the \textsc{Formulator} output is incorrect, it will generate a JSON representation $\mathcal{R}'$ by itself, and the framework will use this $\mathcal{R}'$ to enter \textsc{Code Generator} again. We set the maximum number of loops to be 5.

\subsection{Choice of Solver}
\label{sec:method_why_solver}
As a framework that formulates and solves planning problems as optimization problems, \ours can be adapted to use any planner or solver by modifying the requirements in prompt to follow the syntax of new solvers. In this work, we compare with popular PDDL and MILP solvers and choose SMT solver with following reasons: SMT allows explicit goal and constraint assertion from scratch, which can solve both single-step multi-constraint problems and multi-step problems, while PDDL solvers require rigidly formatted PDDL domain and problem files, which limits its applicability for non-PDDL problems. SMT is complete and sound, and can guarantee optimal plans when used for optimization-based planning. In contrast, many PDDL planners in practice prioritize speed and may use incomplete search strategies by default, though some do ensure completeness. Additionly, for all optimization solvers like SMT and MILP, building optimization problems involves the same steps: defining the goal, constraints, and decision variables, and writing codes to encode relationships between decision variables. Thus, utilizing any optimization planner has a similar process. We show how easily our framework could adapt to use MILP by including prompt differences and output examples in Appendix~\ref{sec:appendix_milp}. We selected SMT over MILP because the SMT Z3 solver is publicly available and more accessible to all users than the Gurobi MILP solver, which requires licenses and limits the number of devices per license.

\section{Experimental Results}
\label{sec:experiments}

\subsection{Domains}
We test on 9 planning problems, which includes 5 multi-constraint decision making tasks,\textbf{Coffee}, \textbf{Workforce}, \textbf{Facility}, \textbf{Task Allocation}, and \textbf{Warehouse}, and 4 multi-step tasks, \textbf{Blocksworld}, \textbf{Mystery Blocksworld}, \textbf{Movie}, and \textbf{Gripper}~\citep{li2023large, valmeekam2024planbench, stein2023autoplanbench}. Task descriptions and complexity analysis are included in Appendix~\ref{sec:domains}.
The queries are either what-if questions that change/add constraints to the existing scenarios or different task initial and goal conditions. Task inputs including example queries are given in Appendix~\ref{sec:appendix_example}.

\subsection{\ours Performance}
\label{sec:exp-main}
We evaluate \ours on 9 tasks with GPT-4o~\citep{gpt-4o} and Claude 3.5 Sonnet~\citep{claude3.5} with temperature 0. Each task comes with natural language task descriptions, queries, background information, APIs. \ours takes these inputs and outputs plans with no task-specific examples. We use optimal rate as the evaluation metric, measuring whether plans are optimal for given the task and query. We also include success rate as another metric and results in~\ref{sec:success}. 

\textbf{Baselines} We compare \ours against 1) Direct: LLM direct plan generation, 2) CoT: chain-of-thought prompting~\citep{wei2022chain} by asking LLMs to reason before generating the final answer, 3) Code: prompts LLM to generate Python codes to solve the problem, allowing the use of any package or solver, \blue{and 4) Code\_SMT: prompts LLM to generate Python codes using Z3 SMT solver, the same tool we use in LLMFP}. For all baselines, we use both GPT-4o and Claude 3.5 Sonnet and also include a direct plan generation baseline with OpenAI o1-preview~\citep{o1preview}. All baselines are zero-shot with no task-specific examples. All baselines have the same input information as \ours, including task description, task background information or info collection API, and query. \blue{We also provide all baselines with formatters to convert their generated plans to fixed formats for better evaluation.} Please refer to Sec.~\ref{sec:appendix_prompt} for prompts of baselines.

\begin{table*}[t]
\caption{Optimal rate (\%) comparison of \ours with baselines on 5 multi-constraint problems.}
\label{main_5}
\begin{center}
\begin{small}
\begin{tabular}{lccccc|c}
\toprule
Method & Coffee & Workforce & Facility & Task Allocation & Warehouse & Average \\
\midrule
Direct$_{\textsc{GPT-4o}}$ & 0.8 & 2.6 & 0.0 & 0.0 & 0.0 & 0.7 \\
Direct$_{\textsc{o1-preview}}$ & 25.9 & 47.6 & 4.8 & 4.0 & 66.0 & 29.7 \\
CoT$_{\textsc{GPT-4o}}$ & 0.0 & 5.6 & 0.0 & 0.0 & 16.0 & 4.3 \\
Code$_{\textsc{GPT-4o}}$ & 17.7 & 75.8 & 53.9 & 0.0 & 8.0 & 31.1 \\
\blue{Code\_SMT$_{\textsc{GPT-4o}}$} & \blue{0.0} & \blue{10.8} & \blue{0.6} & \blue{0.0} & \blue{2.0} & \blue{2.7} \\
\ours$_{\textsc{GPT-4o}}$ & \textbf{64.7} & \textbf{92.2} & \textbf{70.7} & \textbf{96.0} & \textbf{72.0} & \textbf{79.1} \\
\midrule
Direct$_{\textsc{Claude 3.5 Sonnet}}$ & 0.0 & 0.0 & 0.0 & 0.0 & 0.0 & 0.0 \\
CoT$_{\textsc{Claude 3.5 Sonnet}}$ & 7.1 & 0.0 & 0.0 & 0.0 & 14.0 & 4.2 \\
Code$_{\textsc{Claude 3.5 Sonnet}}$ & 59.8 & 71.9 & 47.3 & 0.0 & 42.0 & 44.2 \\
\blue{Code\_SMT$_{\textsc{Claude 3.5 Sonnet}}$} & \blue{75.6} & \blue{36.8} & \blue{\textbf{49.7}} & \blue{86.0} & \blue{64.0} & \blue{62.4} \\
\ours$_{\textsc{Claude 3.5 Sonnet}}$ & \textbf{80.5} & \textbf{88.7} & 48.2 & \textbf{96.0} & \textbf{90.0} & \textbf{80.7} \\
\bottomrule
\end{tabular}
\end{small}
\end{center}
\end{table*}

\begin{table*}[t]
\caption{Optimal rate (\%) comparison of \ours with baselines on 4 multi-step problems.}
\label{main_4}
\begin{center}
\begin{small}
\begin{tabular}{lcccc|c}
\toprule
Method & Blocksworld & Mystery Blocksworld & Movie & Gripper & Average \\
\midrule
Direct$_{\textsc{GPT-4o}}$ & 41.5 & 0.8 & 85.7 & 0.0 & 32.0 \\
Direct$_{\textsc{o1-preview}}$ & 88.4 & 31.9 & \textbf{100.0} & 52.0 & 68.1 \\
CoT$_{\textsc{GPT-4o}}$ & 39.9 & 2.7 & 81.0 & 0.0 & 30.9 \\
Code$_{\textsc{GPT-4o}}$ & 0.0 & 0.3 & 0.0 & 0.0 & 0.1 \\
\blue{Code\_SMT$_{\textsc{GPT-4o}}$} & \blue{0.0} & \blue{0.0} & \blue{0.0} & \blue{4.0} & \blue{1.0}  \\
\ours$_{\textsc{GPT-4o}}$ & \textbf{96.2} & \textbf{77.7} & \textbf{100.0} & \textbf{76.0} & \textbf{87.5} \\
\midrule
Direct$_{\textsc{Claude 3.5 Sonnet}}$ & 43.2 & 0.5 & \textbf{100.0} & 12.0 & 38.9 \\
CoT$_{\textsc{Claude 3.5 Sonnet}}$ & 52.8 & 2.8 & \textbf{100.0} & 28.0 & 45.9 \\
Code$_{\textsc{Claude 3.5 Sonnet}}$ & 0.0 & 0.0 & 0.0 & 0.0 & 0.0 \\
\blue{Code\_SMT$_{\textsc{Claude 3.5 Sonnet}}$} & \blue{0.0} & \blue{0.0} & \blue{0.0} & \blue{0.0} & \blue{0.0} \\
\ours$_{\textsc{Claude 3.5 Sonnet}}$ & \textbf{93.0} & \textbf{98.0} & \textbf{100.0} & \textbf{76.0} & \textbf{91.8} \\
\bottomrule
\end{tabular}
\end{small}
\end{center}
\end{table*}

\textbf{Results and Analysis} We include the optimal rate comparison of \ours and baselines on 5 multi-constraint problems and 4 multi-step problems in Table~\ref{main_5} and~\ref{main_4}. There are \blue{four} key takeaways:

First, \ours achieves strong performance across all 9 tasks, significantly outperforming all baselines. For GPT-4o, \ours achieves an average of 83.7\% optimal rate across 9 tasks (79.1\% for 5 multi-constraint problems and 87.5\% for 4 multi-step problems). For Claude 3.5 Sonnet, \ours achieve an 86.8\% optimal rate across 9 tasks (80.7\% for 5 multi-constraint problems and 91.8\% for 4 multi-step problems). For 5 multi-constraint problems, \ours$_{\textsc{GPT-4o}}$ and \ours$_{\textsc{Claude 3.5 Sonnet}}$ outperform best baselines Code$_{\textsc{GPT-4o}}$ and \blue{Code\_SMT$_{\textsc{Claude 3.5 Sonnet}}$} by a large margin of 48.0\% and \blue{18.3\%}. For 4 multi-step problems, \ours$_{\textsc{GPT-4o}}$ and \ours$_{\textsc{Claude 3.5 Sonnet}}$ outperform Direct$_{\textsc{o1-preview}}$ and CoT$_{\textsc{Claude 3.5 Sonnet}}$  by an average of 19.4\% and 45.9\%. This highlights both the effectiveness and the generalization capability of \ours.

Second, among baselines, Code works better for multi-constraint problems, while Direct and CoT work better for multi-step problems. This validates that the skills required for solving different tasks are different. For multi-constraint problems, as heavy calculations are required to test every possible solution, it is hard for LLMs to directly plan, even with the strongest o1-preview model. For multi-step problems, since Code tries to use a PDDL planner, which requires LLM to generate fixed-format PDDL domain and problem files, it almost always fails to generate and call them correctly. While it is easier for LLMs to directly devise plans as the preconditions and effects of each action are easier to reason about than calculations. This further proves that \ours can tackle problems that are fundamentally different because it uses a universal and formal approach for all tasks. 

Third, for Direct and CoT, Mystery Blocksworld's performance degrades largely compared to Blocksworld, though they are fundamentally same problems. Changing predicate and action names to illogical names makes LLMs hard to understand the problem and generate reasonable plans. However, \ours still can obtain an overall strong optimal rate of 77.7\% and 98.0\% on Mystery Blocksworld for GPT-4o and Claude 3.5 Sonnet. This shows \ours is robust to obfuscated problems, as it can encode the problem as long as the problem is clearly defined regardless of the names.

\blue{Fourth, for multi-constraint problems, Code\_SMT$_{\textsc{Claude 3.5 Sonnet}}$ improves 18.2\% compared to Code$_{\textsc{Claude 3.5 Sonnet}}$, though Code\_SMT$_{\textsc{GPT-4o}}$ performs poorly. This showcases the strong coding capability of Claude 3.5 Sonnet, especially the capability to understand and utilize the SMT solver. At the same time, it showcases the instability for different LLMs to reach strong performance, motivating the need of frameworks like \ours to overcome existing limitations of LLMs.} 

To summarize, \ours is capable of solving all 9 tasks with strong performance and is robust to fundamentally different and obfuscated problems. We show the performance of \ours across iterations, time and cost statistics, and failure analysis in Appendix~\ref{sec:iter}, ~\ref{sec:appendix_timecost}, and~\ref{sec:appendix_failure}.

\subsection{Effectiveness of \ours Components}
\label{sec:exp-ablation}
We then validate each component of \ours with ablation experiments on 9 tasks. We examine the effectiveness of \textsc{Definer}, \textsc{Formulator}, and \textsc{Self Assess \& Modification} by removing these components from our framework one at a time and comparing with \ours. We do not remove \textsc{Code Generator} and \textsc{Result Formatter} because they are the necessary components of \ours to deliver outputs. We use GPT-4o as the LLMs and optimal rate as the evaluation metric. 

\textbf{Results and Analysis}
We include the optimal rate performance comparison of \ours and baselines on 9 problems in Table~\ref{ablation}. 
From Table~\ref{ablation} there are two key takeaways: 

First, removing any of the 3 components from \ours negatively affects the performance. For multi-constraint problems, removing \textsc{Definer}, \textsc{Formulator}, and \textsc{Self Assess \& Modification} lowers the optimal rate by 15.4\%, 22.2\%, and 21.9\%. For multi-step problems, removing \textsc{Formulator} and \textsc{Self Assess \& Modification} reduces the optimal rate by 87.4\% and 12.4\%. 

Second, for different problems, the most effective components are different. Coffee degrades the most for No \textsc{Definer}; Warehouse and all multi-step problems drop the most for No \textsc{Formulator}; and Workforce decreases the most for No \textsc{Self Assess \& Modification}. This again validates the diversity of the 9 problems and how they require different efforts to be successfully solved. Thus, \ours is an overall framework that could aid the process of planning from all aspects. 


To summarize, all three components in \ours are effective and could account for diverse problems by providing comprehensive aids to solve planning problems.  

\begin{table*}[t]
\caption{Optimal rate (\%) comparison when removing some key components of \ours on all 9 tasks. LLMs used are GPT-4o.}
\label{ablation}
\begin{center}
\begin{small}
\begin{tabular}{lcccc}
\toprule
\multicolumn{1}{c}{Domain} & 
\multicolumn{1}{c}{No Definer} & \multicolumn{1}{c}{No Formulator} & \multicolumn{1}{c}{No Self Assess \& Modification} & \multicolumn{1}{c}{\ours}\\

\midrule
\multicolumn{1}{c}{Coffee} & 
\multicolumn{1}{c}{8.6} & \multicolumn{1}{c}{56.4} & \multicolumn{1}{c}{55.3} & \multicolumn{1}{c}{\textbf{64.7}}\\

\multicolumn{1}{c}{Workforce} & 
\multicolumn{1}{c}{84.4} & \multicolumn{1}{c}{80.5} & \multicolumn{1}{c}{27.3} & \multicolumn{1}{c}{\textbf{92.2}}\\

\multicolumn{1}{c}{Facility} & 
\multicolumn{1}{c}{61.6} & \multicolumn{1}{c}{53.7} & \multicolumn{1}{c}{53.7} & \multicolumn{1}{c}{\textbf{70.7}}\\

\multicolumn{1}{c}{Task Allocation} & 
\multicolumn{1}{c}{74.0} & \multicolumn{1}{c}{92.0} & \multicolumn{1}{c}{96.0} & \multicolumn{1}{c}{\textbf{96.0}}\\

\multicolumn{1}{c}{Warehouse} & 
\multicolumn{1}{c}{\textbf{90}} & \multicolumn{1}{c}{2.0} & \multicolumn{1}{c}{54.0} & \multicolumn{1}{c}{72.0}\\

\midrule
\multicolumn{1}{c}{Average} & 
\multicolumn{1}{c}{63.7} & \multicolumn{1}{c}{56.9} & \multicolumn{1}{c}{57.2} & \multicolumn{1}{c}{\textbf{79.1}}\\
\midrule
\midrule

\multicolumn{1}{c}{Blocksworld} & 
\multicolumn{1}{c}{N/A} & \multicolumn{1}{c}{0.2} & \multicolumn{1}{c}{95.3} & \multicolumn{1}{c}{\textbf{96.2}}\\

\multicolumn{1}{c}{Mystery Blocksworld} & 
\multicolumn{1}{c}{N/A} & \multicolumn{1}{c}{0.0} & \multicolumn{1}{c}{74.4} & \multicolumn{1}{c}{\textbf{77.7}}\\

\multicolumn{1}{c}{Movie} & 
\multicolumn{1}{c}{N/A} & \multicolumn{1}{c}{0.0} & \multicolumn{1}{c}{66.7} & \multicolumn{1}{c}{\textbf{100.0}}\\

\multicolumn{1}{c}{Gripper} & 
\multicolumn{1}{c}{N/A} & \multicolumn{1}{c}{0.0} & \multicolumn{1}{c}{64.0} & \multicolumn{1}{c}{\textbf{76.0}}\\

\midrule
\multicolumn{1}{c}{Average} & 
\multicolumn{1}{c}{N/A} & \multicolumn{1}{c}{0.1} & \multicolumn{1}{c}{75.1} & \multicolumn{1}{c}{\textbf{87.5}}\\

\bottomrule
\end{tabular}
\end{small}
\end{center}
\end{table*}

\subsection{\ours with Task-Specific Example}
Although \ours is capable of achieving strong performance on a wide range of problems with no task-specific example, we test \ours by only replacing the two examples in \textsc{Formulator} to one task-specific example on Coffee task to see how much the task-specific example could further improve \ours. Queries of Coffee tasks are what-if questions and are categorized into 7 sets. Each set is a type of question. For example, every query in Set 1 asks about what if the demand in some cafes increases some amount. For each set, we include one task-specific example in \textsc{Formulator} prompt, test \ours$_{\textsc{task-specific}}$ over this set, and compare with \ours over this set. We use GPT-4o for LLMs. From the results in Table~\ref{incontext}, we observe that on average, \ours$_{\textsc{task-specific}}$ improves the performance of \ours by 24.2\%. The performance of Set 3 increases the most. This set asks queries like \textit{``What led to the decision to use supplier3 for the roasting facility at roastery1?"}. It is both plausible to test \textit{``using supplier3"} or to test \textit{``not using supplier3"} to answer the question. However, the ground truth answer for this type of question is to \textit{``not using supplier 3"}. They are confusing queries even for humans to understand, let alone LLMs. Thus, for these queries, adding task-specific examples significantly improves the performance. To summarize, \ours can achieve strong performance with no task-specific example, but easily adding task-specific examples only to \textsc{Formulator} improves the performance, especially when the task or query is not clearly presented.

\begin{table*}[t]
\caption{Optimal rate (\%) comparison of \ours and \ours with one task-specific example in Formulator on \textbf{Coffee} task. Sets represent different types of what-if questions. LLMs are GPT-4o.}
\label{incontext}
\begin{center}
\begin{small}
\begin{tabular}{lccccccc|c}
\toprule
Method & 
Set 1 & Set 2 & Set 3 & Set 4 & Set 5 &
Set 6 & Set 7 &
Average\\
\midrule
\ours & 58.3
& 70.9 & 11.8 & 42.4 & 80.0 & 83.3 & 81.5 & 61.2\\
\ours$_{\textsc{task-specific}}$ & 78.3
& 72.7 & 70.6& 84.8 & 91.4 &100.0 & 100.0& 85.4 \\

\bottomrule
\end{tabular}
\end{small}
\end{center}
\end{table*}

\section{Conclusion}
\label{sec:conclusion}
To account for the challenging trade-off between flexibility and task complexity for LLM planning, we observe that the core of many planning problems lies in optimization problems and propose a universal approach for LLMs to solve planning problems. We propose \ours, a general-purpose LLM-based framework that captures key information from planning problems and formally formulates and solves them as optimization problems, with no task-specific examples needed. We test \ours on 9 diverse planning tasks with two LLMs, proving \ours can achieve strong performance over fundamentally different tasks and showing the effectiveness of components in \ours.

\textbf{Limitations} \ours needs clear and detailed task descriptions and queries. It is hard for \ours to define the problems' goals and constraints if the task description is ambiguous or missing some important information. In addition, since \ours solves encoded planning problems with optimization solvers, the capability of \ours depends on the strength of the solvers. For massive databases with numerous feasible plans, the speed for solvers to search for optimal plans is slow. Ways to mitigate this is to introduce heuristics to prioritize a portion of the choices or to switch from solving optimization problems to satisfaction problems for planning tasks that do not require optimality. 
\section{Acknowledgments}
This work was supported by ONR under Award N00014-22-1-2478 and MIT-IBM Watson AI Lab. However, this article solely reflects the opinions and conclusions of its authors.

\bibliography{iclr2025_conference}
\bibliographystyle{iclr2025_conference}

\clearpage
\onecolumn
\newpage
\addtocontents{toc}{\protect\setcounter{tocdepth}{2}}
\renewcommand{\contentsname}{Planning Anything with Rigor: General-Purpose Zero-Shot Planning with LLM-based Formalized Programming}
\tableofcontents 
\appendix
\section{Appendix}

\subsection{Domains \blue{and Complexity Analysis}}
\label{sec:domains}
We test on 9 planning problems, including 5 multi-constraint decision making tasks and 4 multi-step tasks~\citep{li2023large, valmeekam2024planbench, stein2023autoplanbench}:
\begin{itemize}[left=0pt]
  \item \textbf{Coffee} Coffee company sources beans from three suppliers with fixed capacity, roasts them at two facilities into dark or light coffee, and ships the roasted coffee to three retail locations. The company aims to minimize the total shipping and roasting cost while fulfilling the demand at each retail location. There are 266 different queries of 7 types in the dataset. 
  \item \textbf{Workforce} Assign workers to shifts; each worker may or may not be available on a particular day. The goal is to minimize the total payments to workers while fulfilling the shift requirements for two weeks. There are 231 different queries of 5 types in the dataset. 
  \item \textbf{Facility} A company currently ships its product from 5 plants to 4 warehouses. It is considering closing some plants to reduce costs. The goal is to decide which plant(s) to close to minimize transportation and fixed costs. There are 165 different queries of 4 types in the dataset. 
  \item \textbf{Task Allocation} Given tasks and three robots skilled in different tasks, the goal is to assign tasks to robots to minimize finish time. The finish time counts when the last robot stops working. There are 50 different queries describing the number of different tasks. This task and its data and queries are created by us.
  \item \textbf{Warehouse} The robots need to finish tasks by visiting stations that are capable of accomplishing corresponding tasks. The goal is to find the list of stations while minimizing the total distance traveled. There are 50 different queries that include the random-length list of tasks to finish. This task and its data and queries are created by us.
  \item \textbf{Blocksworld} The robot has four actions: pickup, putdown, stack, and unstack. The goal is to stack the blocks  in the scene from their initial setup to a specific order with minimum steps. There are 602 different queries that describe blocks' initial conditions and goal states.
  \item \textbf{Mystery Blocksworld} An obfuscated version of Blocksworld. The action and predicate names are replaced with names that logically make no sense. There are 602 different queries that describe objects' initial conditions and goal states.
  \item \textbf{Movie} The goal is to get the required snacks, watch the movie, and recover the movie and counter to the original state with minimum steps. There are 21 different queries that describe objects' initial conditions and goal states.
  \item \textbf{Gripper} There are robots and balls in different rooms. Each robot, with two grippers, can pick, drop, and move balls between rooms. The goal is to place balls in specific rooms with minimum steps. There are 25 different queries describing objects' initial conditions and goal states.
\end{itemize}
The queries for Coffee, Workforce, and Facility are what-if questions that change or add constraints to the existing scenarios. The queries of the rest tasks are different task initial and goal conditions. Task inputs including example queries are given in Appendix~\ref{sec:appendix_example}.

\textbf{Mathematical Representation}
\blue{We use the benchmark from~\citep{li2023large} for the first 3 problems (Coffee, Workforce, and Facility), in which these 3 problems are built as Mixed-integer linear programming (MILP) problems. As an example, here is the problem definition of Coffee as a MILP problem (Defined in Page 6-7 from~\citep{li2023large}):}

\blue{$x_{s,r}$ is the number of units purchased from supplier $s$ for roasting facility $r$, and $y^{L}_{r,\ell}$ and $y^{D}_{r,\ell}$ is the amount of light and dark roast sent to retail location $\ell$ from roasting facility $r$. $C_s$ is the capacity for each supplier $s$, and $D^{L}_{\ell}$ and $D^{D}_{\ell}$ are demand for light and dark roast for each retail location $\ell$. There is a cost $c_{s,r}$ for each unit purchased from supplier $s$ for roasting facility $r$, a shipping cost of $g_{r,\ell}$ for each unit sent to retail location $\ell$ from roasting facility $r$, and a roasting cost $h^{L}_{r}$ and $h^{D}_{r}$ per unit of light roast and dark roast respectively in facility $r$.
\begin{align*}
    \text{minimize} & \quad \left( \sum_{s,r} x_{s,r} \cdot c_{s,r} + \sum_{r,\ell} y^{L}_{r,\ell} \cdot h^{L}_{r,\ell} +\sum_{r,\ell} y^{D}_{r,\ell} \cdot h^{D}_{r} + \sum_{r,\ell} \left( y^{L}_{r,\ell} + y^{D}_{r,\ell} \right) \cdot g_{r,\ell} \right) \\
    \text{subject to} & \\
    & \sum_{r} x_{s,r} \leq C_{s}, \quad \forall s \quad \text{(Supplier capacity constraint)} \\
    & \sum_{s} x_{s,r} = \sum_{\ell} \left( y^{L}_{r,\ell} + y^{D}_{r,\ell} \right), \quad \forall r \quad \text{(Conservation of flow constraint)} \\
    & \sum_{r} y^{L}_{r,\ell} \geq D^{L}_{\ell}, \quad \forall \ell \quad \text{(Light coffee demand constraint)} \\
    & \sum_{r} y^{D}_{r,\ell} \geq D^{D}_{\ell}, \quad \forall \ell \quad \text{(Dark coffee demand constraint)} \\
    & x_{s,r}, y^{L}_{r,\ell}, y^{D}_{r,\ell} \in \mathbb{Z}^{+}, \quad \forall s, r, \ell \quad \text{(Integrality constraint)}
\end{align*}}   

\textbf{Complexity Analysis}

\blue{The Coffee problem can be framed as a max-flow problem, which can be solved in polynomial time. Specifically, some algorithms can solve the max-flow problem with $O(VE)$ or $O(V^2E)$}

\blue{The Workforce problem, with no additional constraint, can also be framed as a max-flow problem. However, different types of constraints are added by users to form different instances. Some types of queries can increase the complexity. For example, "What if Gu and Bob cannot work on the same day?". Adding constraints to introduce conflicting workers turns the problem to be as hard as a maximum independent set problem(also NP-Hard), where we add an edge between conflicted workers, and finding the maximum indentpendent set.}

\blue{The facility problem is a NP-Hard problem Capacited Facility Location Problem(CFLP). The formal definition of CFLP is as below: 
\begin{align*}
\min \sum_{i=1}^n \sum_{j=1}^m c_{ij} d_j y_{ij} + \sum_{i=1}^n f_i x_i \\
\text{s.t.} \quad \sum_{i=1}^n y_{ij} = 1 \quad \text{for all } j = 1, \ldots, m \\
\sum_{j=1}^m d_j y_{ij} \leq u_i x_i \quad \text{for all } i = 1, \ldots, n \\
y_{ij} \geq 0 \quad \text{for all } i = 1, \ldots, n \text{ and } j = 1, \ldots, m \\
x_i \in \{0, 1\} \quad \text{for all } i = 1, \ldots, n
\end{align*}
where $x_i = 1$ if facility $i$ is open, and $x_i = 0$ otherwise. $y_{ij}$ for $i = 1, \ldots, n$ and $j = 1, \ldots, m$, which represents the fraction of the demand $d_j$ filled by facility $i$.}

\blue{For the Task Allocation problem, since it is equivalent to a multi-agent traveling salesman problem(agent=robots, tasks=cities), it reduces from the classic traveling salesman problem (TSP) and thus is also NP-hard. }

\blue{For the Warehouse problem, as TSP is a special case when one station can be used to finish one specific task and there are no extra stations, the Warehouse problem is at least as complex as TSP, thus is also NP-hard.}

\blue{For multi-step problems, Blocksworld is proved to be a NP-hard problem~\citep{gupta1992complexity}, so same for Mystery Blocksworld as it is the same problem with obfuscated names. Although there is no existing proof, Movie has 13 predicates and 9 possible actions, and Gripper has 4 types of objects (rooms, objects, robots, grippers), 4 predicates, and 3 possible actions. These show that they are not simple straightforward tasks.
}

\textbf{\ours Performance on Sokoban} \blue{To further test capability of LLMFP on even more challenging tasks, we tested LLMFP on the Sokoban environment, a NP-Hard problem with large maps thus needs more variables. We created an evaluation set containing 15 queries describing the game setup and goals with different map sizes and number of boxes. We have five queries with 5x5 maps and 1 box, five queries with 6x6 maps and 1 box, and five queries with 5x5 maps and 2 boxes. The evaluation results are presented in the following table.
}
\begin{table*}[!ht]
\caption{\blue{Optimal rates (\%) comparison of \ours with baselines on Sokoban task.}}
\label{sokoban}
\begin{center}
\begin{small}
\begin{tabular}{cccccc}
\toprule
Direct$_{\textsc{GPT-4o}}$ & Direct$_{\textsc{o1-preview}}$ & CoT$_{\textsc{GPT-4o}}$ & Code$_{\textsc{GPT-4o}}$ & Code\_SMT$_{\textsc{GPT-4o}}$ & \ours$_{\textsc{GPT-4o}}$ \\
\midrule
0.0 & 26.7 & 0.0 & 0.0 & 0.0 & 80.0\\
\bottomrule
\end{tabular}
\end{small}
\end{center}
\end{table*}

\blue{As can be observed, LLMFP achieves a success rate of 80\%, outperforming the baselines. The new results, along with other problems, showcase the potential of LLMFP to solve complex problems.}

\blue{The major failure mode is: when the generated codes initialize the adjacent predicate, it only initializes adjacent positions to be True but fails to initialize unmentioned positions to be False (since the query only mentions position\_x and position\_y are adjacent), so the solver would set the non-adjacent positions to adjacent to get solution with fewer steps. In addition, although LLMFP is demonstrated to be capable of correctly encoding and solving the Sokoban problem, it is true that there are many more variables in the Sokoban problem than in other tasks because the problem is represented with a map with a large number of different positions. This slows down the speed of the SMT solver. To mitigate this problem, some potential solutions include 1) introducing methods to estimate the lower and upper bounds of step numbers needed and start from there, 2) developing heuristics to prioritize some possible options first, and 3) developing methods that put attention on a part of the map and ignore the unnecessary positions in the map. We would love to extend our work to explore these directions to make our framework more efficient.
}

\clearpage
\subsection{\ours Performance over Iterations}
\label{sec:iter}

\begin{figure}[!ht]
  \includegraphics[width=\linewidth]{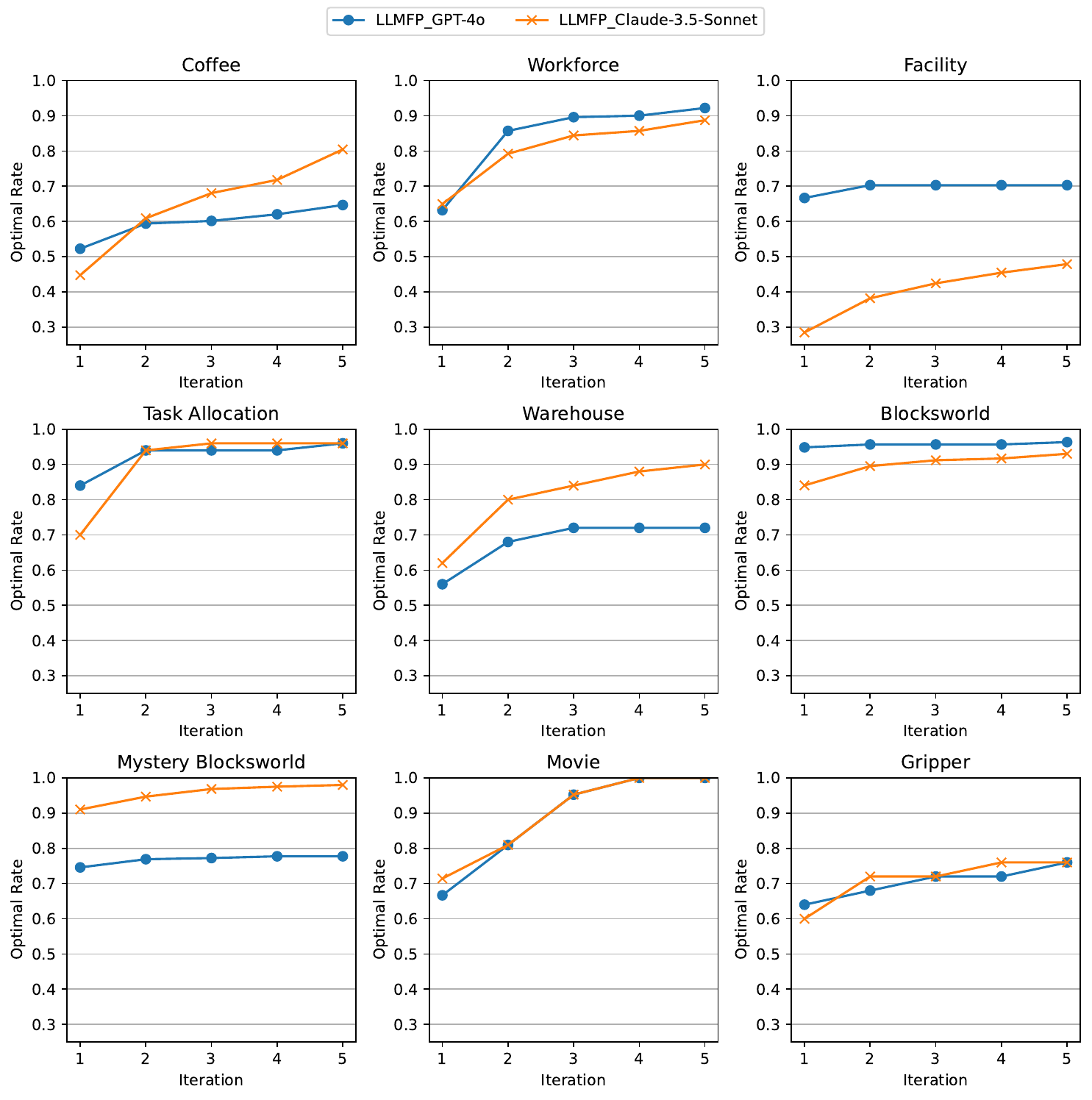} 
  \caption {Optimal rates of models across \ours Iterations}
  \label{fig:iter}
\end{figure}

Fig.~\ref{fig:iter} shows the performance of \ours over 5 iterations. The key observation is: number of iterations of Self Assess \& Midification stage enables \ours to further improve the optimal rates, although we can observe that \ours does not need extensive iterations to achieve an overall satisfying performance. 

\clearpage
\subsection{\blue{Additional Metric Performance: Success rate}}
\label{sec:success}
\blue{In addition to the optimal rate, we also include another metric, success rate to evaluate baselines and \ours. We include the result in Table~\ref{success_5} and Table~\ref{success_4}.}

\blue{Note that although for multi-constraint problems the optimization goal is described in the task description, we exclude the optimization goal when calculating the success rate and only evaluate whether the plan fulfills the task setup and the query. This would largely decrease the difficulty of multi-constraint problems. For example, even assigning all tasks to one robot is considered a success for the task allocation task. Thus, the success rates of all baselines for multi-constraint problems are significantly higher than the optimal rates. However, although the success rates of LLMFP almost remains the same as optimal rates since the SMT solver guarantees to output the optimal result with correct encoding, the performance of LLMFP still outperforms other baselines, with an average of 86.4\%, 18.1\% higher than the best baseline. }

\blue{While for the multi-step problems, considering all initial conditions, predicate and action definitions, and goals are the same, developing a reasonable and correct plan is not significantly easier than developing an optimal plan with the least number of steps. Thus, the success rates of baselines are improved, but not significantly, compared to the optimal rates.}

\begin{table*}[htb!]
\caption{\blue{Success rate (\%) comparison of \ours with baselines on 5 multi-constraint problems.}}
\label{success_5}
\begin{center}
\begin{small}
\begin{tabular}{lccccc|c}
\toprule
Method & Coffee & Workforce & Facility & Task Allocation & Warehouse & Average \\
\midrule
Direct$_{\textsc{GPT-4o}}$ & 5.6 &	54.5 & 31.7 & \textbf{100.0} & 42.0 & 46.8 \\
Direct$_{\textsc{o1-preview}}$ & 26.3 & \textbf{92.6} & 41.5 & 94.0 & 86.0 & 68.1 \\
CoT$_{\textsc{GPT-4o}}$ & 17.7 & 72.3 & 31.7 & \textbf{100.0} & 82.0 & 60.7 \\
Code$_{\textsc{GPT-4o}}$ & 18.8 & 76.2 & 64.6 & 92.0 & 90.0 & 68.3 \\
Code\_SMT$_{\textsc{GPT-4o}}$ & 0.0 & 10.8 & 1.2 & 0.0 & 34.0 & 9.2 \\
\ours$_{\textsc{GPT-4o}}$ & \textbf{64.7} & 92.2 & \textbf{79.3} & \textbf{100.0} & \textbf{96.0} & \textbf{86.4} \\
\midrule
Direct$_{\textsc{Claude 3.5 Sonnet}}$ & 5.3 & \textbf{91.3} & 36.0 & \textbf{100.0} & 76.0 & 61.7 \\
CoT$_{\textsc{Claude 3.5 Sonnet}}$ & 10.9 & 60.6 & 1.2 & \textbf{100.0} & 96.0 & 53.7 \\
Code$_{\textsc{Claude 3.5 Sonnet}}$ & 61.3 & 89.2 & 59.1 & \textbf{100.0} & 60.0 & 73.9 \\
Code$_{\textsc{Claude 3.5 Sonnet}}$ & 77.1 & 39.0 & 59.1 & 90.0 & 74.0 & 67.8 \\
\ours$_{\textsc{Claude 3.5 Sonnet}}$ & \textbf{80.5} & 88.7 & \textbf{61.6} & \textbf{100.0} & \textbf{92.0} & \textbf{84.6} \\
\bottomrule
\end{tabular}
\end{small}
\end{center}
\end{table*}

\begin{table*}[htb!]
\caption{\blue{Success rate (\%) comparison of \ours with baselines on 4 multi-step problems.}}
\label{success_4}
\begin{center}
\begin{small}
\begin{tabular}{lcccc|c}
\toprule
Method & Blocksworld & Mystery Blocksworld & Movie & Gripper & Average \\
\midrule
Direct$_{\textsc{GPT-4o}}$ & 56.1 & 1.0 & 90.5 & 16.0 & 40.9\\
Direct$_{\textsc{o1-preview}}$ & 90.9 & 37.9 & \textbf{100.0} & \textbf{76.0} & 76.2 \\
CoT$_{\textsc{GPT-4o}}$ & 62.0 & 3.0 & 95.2 & 10.0 & 42.5 \\
Code$_{\textsc{GPT-4o}}$ & 0.0 & 0.3 & 0.0 & 0.0 & 0.1 \\
Code\_SMT$_{\textsc{GPT-4o}}$ & 0.2 & 0.0 & 0.0 & 4.0 & 1.0 \\
\ours$_{\textsc{GPT-4o}}$ & \textbf{96.2} & \textbf{77.7} & \textbf{100.0} & \textbf{76.0} & \textbf{87.5} \\
\midrule
Direct$_{\textsc{Claude 3.5 Sonnet}}$ & 54.5 & 0.5 & \textbf{100.0} & 56.0 & 52.7 \\
CoT$_{\textsc{Claude 3.5 Sonnet}}$ & 76.1 & 3.2 & \textbf{100.0} & 72.0 & 62.8 \\
Code$_{\textsc{Claude 3.5 Sonnet}}$ & 0.0 & 0.0 &  0.0 & 0.0 & 0.0 \\
Code\_SMT$_{\textsc{Claude 3.5 Sonnet}}$ & 0.0 & 0.0 & 4.0 & 0.0 & 1.0 \\
\ours$_{\textsc{Claude 3.5 Sonnet}}$ & \textbf{93.4} & \textbf{98.0} & \textbf{100.0} & \textbf{76.0} & \textbf{91.8} \\
\bottomrule
\end{tabular}
\end{small}
\end{center}
\end{table*}

\clearpage
\subsection{\blue{Time and Cost Statistics and Analysis}}
\label{sec:appendix_timecost}
\blue{Table~\ref{time_5} and Tabel~\ref{time_4} show the wall time comparison of all methods for GPT-4o on 9 tasks. From the results, we could observe that the time taken for \ours, although longer than most of the baselines, is within reasonable ranges. Especially, for multi-constraint problems, it is shorter than Direct with o1-preview because of the inherent difficulty for LLMs to solve these combinatorial optimization problems.}

Table~\ref{time} shows the detailed time statistics of all components of \ours for GPT-4o on 9 tasks. We could observe that for both LLM querying time and solver running time, all stages of \ours requires reasonable runtime. The longest runtime is prompting \textsc{Formulator} it is time-consuming to reason about all needed variables and information to form representation formulation. 

\begin{table*}[htb!]
\caption{\blue{Average wall time (s) per query comparison for 5 multi-constraint problems with GPT-4o.}}
\label{time_5}
\begin{center}
\begin{small}
\begin{tabular}{lccccc|c}
\toprule
Method & Coffee & Workforce & Facility & Task Allocation & Warehouse & Average \\
\midrule
Direct$_{\textsc{GPT-4o}}$ & 8.8 &	2.2 & 2.1 & 1.8 & 0.9 & 3.2 \\
Direct$_{\textsc{o1-preview}}$ & 104.2&	63.9&	77.7&	70.5&	63.7&	76.0 \\
CoT$_{\textsc{GPT-4o}}$ & 16.9&	12.0&	6.0&	9.6&	7.4&	10.4 \\
Code$_{\textsc{GPT-4o}}$ & 30.6&	10.0&	8.2&	5.7&	7.1&	12.3 \\
Code\_SMT$_{\textsc{GPT-4o}}$ & 30.0&	15.3&	10.3&	15.0&	8.3&	15.8 \\
\ours$_{\textsc{GPT-4o}}$ & 87.1&	55.1&	29.9&	62.3&	28.9&	52.7 \\
\bottomrule
\end{tabular}
\end{small}
\end{center}
\end{table*}

\begin{table*}[htb!]
\caption{\blue{Average wall time (s) per query comparison for 4 multi-step problems with GPT-4o.}}
\label{time_4}
\begin{center}
\begin{small}
\begin{tabular}{lcccc|c}
\toprule
Method & Blocksworld & Mystery Blocksworld & Movie & Gripper & Average \\
\midrule
Direct$_{\textsc{GPT-4o}}$ & 0.7&	0.7&	0.5&	8.8&	2.7\\
Direct$_{\textsc{o1-preview}}$ & 26.3&	87.9&	25.7&	23.8&	40.9 \\
CoT$_{\textsc{GPT-4o}}$ & 2.1&	4.0&	1.0&	10.2&	4.3\\
Code$_{\textsc{GPT-4o}}$ & 19.7&	8.9&	7.3&	8.2&	11.0\\
Code\_SMT$_{\textsc{GPT-4o}}$ & 9.1&	8.5&	10.6&	12.9&	10.3\\
\ours$_{\textsc{GPT-4o}}$ & 43.3&	48.3&	58.6&	141.6&	73.0\\
\bottomrule
\end{tabular}
\end{small}
\end{center}
\end{table*}

\begin{table*}[!ht]
\caption{Average time (s) spent per query for all components of \ours$_{\textsc{GPT-4o}}$ on all 9 tasks.}
\label{time}
\begin{center}
\begin{small}
\begin{tabular}{lcccccc}
\toprule
\multicolumn{1}{c}{Domain} & 
\multicolumn{1}{c}{Definer} & \multicolumn{1}{c}{Formulator} &
\multicolumn{1}{c}{Solver} &
\multicolumn{1}{c}{Formatter} & 
\multicolumn{1}{c}{Code Gen.} & 
\multicolumn{1}{c}{Self Assess \& Mod.}\\

\midrule
\multicolumn{1}{c}{Coffee} & 
\multicolumn{1}{c}{5.6} & 
\multicolumn{1}{c}{10.8} & 
\multicolumn{1}{c}{17.1} & 
\multicolumn{1}{c}{0.1} & 
\multicolumn{1}{c}{14.2} & 
\multicolumn{1}{c}{11.3}\\

\multicolumn{1}{c}{Workforce} & 
\multicolumn{1}{c}{3.4} &
\multicolumn{1}{c}{5.1} & 
\multicolumn{1}{c}{8.3} & \multicolumn{1}{c}{11.0} & \multicolumn{1}{c}{3.1} & \multicolumn{1}{c}{7.8}\\

\multicolumn{1}{c}{Facility} & 
\multicolumn{1}{c}{3.8} &
\multicolumn{1}{c}{7.5} & 
\multicolumn{1}{c}{6.4} & \multicolumn{1}{c}{0.7} & \multicolumn{1}{c}{4.3} & \multicolumn{1}{c}{4.0}\\

\multicolumn{1}{c}{Task Allocation} & 
\multicolumn{1}{c}{8.6} &
\multicolumn{1}{c}{23.8} & 
\multicolumn{1}{c}{5.2} & \multicolumn{1}{c}{0.2} & \multicolumn{1}{c}{6.5} & \multicolumn{1}{c}{5.9}\\

\multicolumn{1}{c}{Warehouse} & 
\multicolumn{1}{c}{3.9} &
\multicolumn{1}{c}{3.1} & 
\multicolumn{1}{c}{6.2} & \multicolumn{1}{c}{0.2} & \multicolumn{1}{c}{4.1} & \multicolumn{1}{c}{3.3}\\

\midrule

\multicolumn{1}{c}{Blocksworld} & 
\multicolumn{1}{c}{N/A} &
\multicolumn{1}{c}{21.0} & 
\multicolumn{1}{c}{14.6} & \multicolumn{1}{c}{0.6} & \multicolumn{1}{c}{1.9} & \multicolumn{1}{c}{3.4}\\

\multicolumn{1}{c}{Mys. Blocksworld}  & \multicolumn{1}{c}{N/A}& 
\multicolumn{1}{c}{24.3} &
\multicolumn{1}{c}{14.6} & 
\multicolumn{1}{c}{0.6} & \multicolumn{1}{c}{2.3} & \multicolumn{1}{c}{4.1}\\

\multicolumn{1}{c}{Movie}  & \multicolumn{1}{c}{N/A}& 
\multicolumn{1}{c}{21.2} &
\multicolumn{1}{c}{9.9} & 
\multicolumn{1}{c}{0.3} & \multicolumn{1}{c}{2.1} & \multicolumn{1}{c}{10.6}\\

\multicolumn{1}{c}{Gripper}  & \multicolumn{1}{c}{N/A}& 
\multicolumn{1}{c}{18.3} &
\multicolumn{1}{c}{16.0} & 
\multicolumn{1}{c}{6.9} & \multicolumn{1}{c}{11.2} & \multicolumn{1}{c}{7.0}\\

\bottomrule
\end{tabular}
\end{small}
\end{center}
\end{table*}

\clearpage
\blue{Table~\ref{cost_compare} shows the average cost comparison of all methods on the coffee task, and Table~\ref{cost_1}, and ~\ref{cost_2} shows the cost statistics of \ours over all 9 tasks. We observe that although \ours is more costly than most of the baselines, it is cheaper than Direct with o1-preview with better performance.} In addition, the average cost per query for all 9 tasks is around 0.1 dollar, indicating \ours is not very costly.
\begin{table*}[!ht]
\caption{\blue{Average cost (\$) per query comparison of \ours$_{\textsc{GPT-4o}}$ on the Coffee task.}}
\label{cost_compare}
\begin{center}
\begin{small}
\begin{tabular}{cccccc}
\toprule
Direct$_{\textsc{GPT-4o}}$ & Direct$_{\textsc{o1-preview}}$ & CoT$_{\textsc{GPT-4o}}$ & Code$_{\textsc{GPT-4o}}$ & Code\_SMT$_{\textsc{GPT-4o}}$ & \ours$_{\textsc{GPT-4o}}$ \\
\midrule
0.008 & 0.536 & 0.013 & 0.023 & 0.024 & 0.139\\
\bottomrule
\end{tabular}
\end{small}
\end{center}
\end{table*}

\begin{table*}[!ht]
\caption{Average cost (\$) per query of \ours$_{\textsc{GPT-4o}}$ on 5 multi-constraint problems.}
\label{cost_1}
\begin{center}
\begin{small}
\begin{tabular}{ccccc}
\toprule
Coffee & Workforce & Facility & Task Allocation & Warehouse \\
\midrule
0.139 & 0.140 & 0.083 & 0.081 & 0.085\\
\bottomrule
\end{tabular}
\end{small}
\end{center}
\end{table*}

\begin{table*}[!ht]
\caption{Average cost (\$) per query of \ours$_{\textsc{GPT-4o}}$ on 4 multi-step problems.}
\label{cost_2}
\begin{center}
\begin{small}
\begin{tabular}{cccc}
\toprule
Blocksworld & Mystery Blocksworld & Movie & Gripper \\
\midrule
0.122 & 0.105 & 0.131 & 0.128 \\
\bottomrule
\end{tabular}
\end{small}
\end{center}
\end{table*}

\clearpage
\subsection{\blue{Baselines Failure Case Analysis}}
\blue{Here we describe the major failure cases for the baselines. Please refer to Appendix~\ref{sec:appendix_output} for example outputs.}

\subsubsection{\blue{Direct, Direct$_{\textsc{o1-preview}}$, and CoT}} 
\blue{For multi-constraint tasks, since they involve various constraints, intensive calculations, and numerous possible solutions, LLMs still do not have the capability to directly solve the optimal solution considering all constraints. They either fail to understand or consider some important constraints or fail to optimize the goal. Although utilizing stronger o1-preview or taking advantages of prompting techniques like CoT could result in less mistakes, the major underlying failure reasons are similar.}

\blue{For multi-step tasks, the major failure cases are the failure to deliver reasonable plans considering preconditions and effects of all actions accurately. }

\subsubsection{\blue{Code}}
\blue{For multi-constraint tasks, the major failure cases are 1) failing to consider all necessary constraints, 2) failing to consider or understand the query, and 3) overwriting the given API. }

\blue{For multi-step tasks, the major failure cases are 1) failing to correctly represent the problem, including the problem setup, predicates, and actions, 2) failing to write codes with correct logic or syntax. }

\subsubsection{\blue{Code\_SMT}}
\blue{For multi-constraint tasks, the major failure cases are same as Code.}

\blue{For multi-step tasks, the major failure cases are 1) Poor SMT Utilization: including failing to distinguish And and Implies, to correctly use SMT Array or Function, or to write correct SMT syntax (or Python syntax sometimes), and 2) Poor Problem Understanding: failing to initialize the initial value of unmentioned predicates (eg. when the query says blocks a, b, d are clear, codes also need to initialize c to be not clear), to assert one action per step, or to update unchanged variables for next step}

\subsubsection{\blue{Theoretical Insights}}
\blue{Overall, LLMs are good at understanding the syntax and semantics of planning problems as optimization problems but are not good at solving optimization problems directly. Specifically,
next-token prediction is fundamentally different from deterministic algorithms for optimization. There’s a growing belief that next token prediction cannot truly model human thought and cannot support human-like capabilities of understanding a problem, imagining, curating, and backtracking plans before executing [1-3]. Specifically, the claim that next token predictions are “ill-suited for planning tasks” is supported by works [4-7], which tested the planning capabilities of LLMs on various planning tasks. These works empirically show that in addition to identifying patterns in language and predicting the next word in a sequence, LLMs still can not truly understand a problem and thus do not have the capability to perform intense calculations to optimize for any objectives. Thus, this is a major reason why baselines are not capable of solving the complex planning problems in our paper. However, since LLMFP teaches LLMs to build the optimization problem step by step and calls the external solver to solve for a plan, this bypasses the need to devise a plan by LLMs themselves.}

\blue{To support this claim that LLMs cannot understand and solve an optimization problem, we conduct an experiment on the Coffee task that, instead of using natural language task descriptions as inputs, we directly map this Coffee task to an optimization problem and use the formal mathematical definition of this problem as the inputs to LLMs. Thus, LLMs do not need to understand the problem and find the underlying constraints, as a formal definition is given and could be directly solved. }

\blue{We tested Direct with the most powerful LLM OpenAI o1-preview model on all queries of Coffee, which only achieves an optimal rate of 34.2\%. Compared to its 25.9\% optimal rate with natural language task description, this is not a significant improvement, given all goals and constraints are clearly formally specified in the new setting. This is consistent with the conclusion that LLMs still cannot solve optimization problems by themselves, even given a formal representation. LLMFP enables LLMs to formalize planning problems as optimization problems.  Since SMT solvers can guarantee to return correct answers given correct input, the high optimal and success rate of LLMFP indicates that LLMFP allows LLM to parse the correct syntax and semantics information of a planning problem from its natural language description to a formal mathematical description. Such translation is also non-trivial when no task-specific examples are provided. As shown in our newly added baseline approach Code\_SMT as shown in Table 1 and 2, when we directly ask LLMs to translate and encode the natural language task description in an SMT format, the optimal rate is low, with an average of 2.7\% and 62.4\% for multi-constraint tasks, and 1.0\% and 0.0\% for multi-step tasks across two LLMs GPT-4o and Claude 3.5 Sonnet.}

\clearpage
\subsection{\ours Failure Case Analysis}
\label{sec:appendix_failure}
Here we analyze the major failure cases for all 9 tasks.
\subsubsection{Coffee}
There are two major failure cases for Coffee tasks: 

First, some queries are not clearly presented, indicating ambiguous information. Queries of Coffee tasks are what-if questions and are categorized into 7 sets. Each set is a type of question. We notice that the type \textit{``supply-roastery"} asks queries like \textit{``What led to the decision to use supplier3 for the roasting facility at roastery1?"}. To answer this question, it is both plausible to test \textit{``using supplier3"} or to test \textit{``not using supplier3"} to see the performance. However, the ground truth answer for this type of questions is to \textit{``not using supplier 3"}. As confusing queries even for human, they are hard for LLMs to understand. Thus, for these queries, \ours sometimes generate codes with opposite meanings as what is expected. 

Second, sometimes \ours \textsc{Definer} fails to consider all implicit constraints. The most easily neglectable implicit constraints are 1) the beans roasted in each roastery do not exceed the beans it receives, and 2) the beans ship from each roastery do not exceed the coffee it roasts. When any of the two constraints are missing, to minimize the cost, the model will automatically set the shipped beans or roasted coffee to be 0, assuming the company delivers coffee without sourcing beans or roasting coffee. 

\subsubsection{Workforce}
There are two major failure cases for Coffee tasks: 

First, sometimes \ours fails to understand the queries. Some of the queries asks questions like \textit{`Can Gu transition from Sun14 to Sun7 for work purposes?'}. The meaning is to \textbf{force} Gu to work on Sun7 and take rest on Sun14. However, sometimes \ours builds variables to test both taking and not taking this transition, and returns solutions with less costs. 

Second, sometimes when the solution space is large, it is hard to find the optimal solution within maximum runtime set for solver. We set the maximum solver runtime to be 15 minutes, which is exceeded when solving some hard queries.
\subsubsection{Facility}
Similarly as the first failure case of Coffee, some queries are not clearly presented. The queries are like \textit{``What justifies the opening of plant 0?"}, which is confusing even for humans. Both opening plant 0 and closing plant 0 to report the costs make sense to answer this query. However, the ground truth meaning of this query is to close plant 0.
\subsubsection{Task Allocation}
\ours only fails one query in Task Allocation. The reason is the \textsc{Formulator} generates wrong values for robot finish time.
\subsubsection{Warehouse}
The major failure case for Warehouse is \textsc{Code Generator} overwrites the provided API \verb|get_distance| and provide 1 as the output during Code Generation. Thus, the distance between each station is mistakenly set to be 1. 
\subsubsection{Blocksworld}
One major failure case for Blocksworld is \textsc{Code Generator} fails to initialize the states of predicates correctly and thoroughly. Since the query will only meantion the predicates that are true, for example, block 1 is on block 2, but when initializing, \ours needs to initialize both mentioned states but also unmentioned states that are false. For example, block 2 is not on block 1. However, \textsc{Code Generator} sometimes fails to consider all unmentioned states.
\subsubsection{Mystery Blocksworld}
Similarly as Blocksworld, Mystery Blocksworld has same failure case. For Mystery Blocksworld, since the predicate and action names are not meaningful, more this kind of errors are made by GPT-4o. However, Claude seems to have better reasoning capability to support it from making more these errors. 
\subsubsection{Movie}
There is no failure case for Movie.
\subsubsection{Gripper}
The major failure case for Gripper is when the solver fails to find the solution because there are some code generation errors, the \textsc{Self Assess \& Modification} sometimes would think it is because the timestep is not enough, thus adding another loop within the original loop. However, this would result in the program to execute forever.

\clearpage
\subsection{\blue{Baselines with Explicit Optimal Requirements}}
\label{sec:baselines_explicit}
\blue{
For all methods including LLMFP and baselines, we implicitly mention the goal of each multi-constraint task in the task description. For example, for the Coffee task, the task description \textit{“...The company's objective is to \textbf{minimize} the total cost, including shipping beans, roasting, and shipping roasted coffee, while ensuring that all coffee produced meets or exceeds the demand at each retail location”} implicitly shows the goal is to find the plan that minimizes the total cost.}

\blue{While for multi-step problems, the methods are not explicitly instructed to provide optimal solutions. Since SMT solver guarantees to find the solution if there exists one, it can rigorously show the solution does not exist for smaller timesteps and increase timestep, thus can always find the optimal solution if the formulation and generated codes are correct. This is an advantage of incorporating a complete and sound solver like SMT in our framework. }

\blue{However, to better understand the capabilities of baselines, we modify the prompts to explicitly instruct them to find the optimal solution and re-evaluate them on the 4 multi-step problems. We add \_Opt to the name of the baselines to represent the baselines with explicit optimal instructions. }

\blue{Table~\ref{optimal_baseline} shows the optimal rate of baselines with explicit instruction on finding the optimal plan. Compared with Table~\ref{main_4}, we could observe that some baselines achieve better performance (from average 0.1\% to 16.4\% for Code\_Opt$_{\textsc{GPT-4o}}$, and from average 30.9\% to 36.7\% for CoT\_Opt$_{\textsc{GPT-4o}}$), while some achieve slightly worse performance(average 68.1\% to 67.0\% for Direct\_Opt$_{\textsc{o1-preview}}$). However, despite the changes due to the explicit instruction to find the optimal plan, \ours still could largely outperform all baselines.
}

\begin{table*}[htb!]
\caption{\blue{Optimal rate (\%) comparison of \ours with baselines that explicitly instructed to generate optimal plans on 4 multi-step problems.}}
\label{optimal_baseline}
\begin{center}
\begin{small}
\begin{tabular}{lcccc|c}
\toprule
Method & Blocksworld & Mystery Blocksworld & Movie & Gripper & Average \\
\midrule
Direct\_Opt$_{\textsc{GPT-4o}}$ & 35.2 & 0.8 & \textbf{100.0} & 0.0 & 34.0 \\
Direct\_Opt$_{\textsc{o1-preview}}$ & 80.9 & 39.0 & \textbf{100.0} & 48.0 & 67.0 \\
CoT\_Opt$_{\textsc{GPT-4o}}$ & 33.4 & 2.3 & 95.2 & 16.0 & 36.7 \\
Code\_Opt$_{\textsc{GPT-4o}}$ &0.0&3.8 &  61.9& 0.0 &  16.4  \\
Code\_SMT\_Opt$_{\textsc{GPT-4o}}$ &0.0  &  0.0&  0.0&  0.0&0.0   \\
\ours$_{\textsc{GPT-4o}}$ & \textbf{96.2} & \textbf{77.7} & \textbf{100.0} & \textbf{76.0} & \textbf{87.5} \\
\midrule
Direct\_Opt$_{\textsc{Claude 3.5 Sonnet}}$ & 40.9 & 1.5 & \textbf{100} & 20.0 & 40.6 \\
CoT\_Opt$_{\textsc{Claude 3.5 Sonnet}}$ & 52.5 & 4.5 & \textbf{100} & 20.0 & 44.2 \\
Code\_Opt$_{\textsc{Claude 3.5 Sonnet}}$ & 0.0 &0.0 &0.0  & 0.0 &  0.0\\
Code\_SMT\_Opt$_{\textsc{Claude 3.5 Sonnet}}$ & 0.0 & 0.0 & 0.0 &  0.0& 0.0 \\
\ours$_{\textsc{Claude 3.5 Sonnet}}$ & \textbf{93.0} & \textbf{98.0} & \textbf{100.0} & \textbf{76.0} & \textbf{91.8} \\
\bottomrule
\end{tabular}
\end{small}
\end{center}
\end{table*}

\clearpage

\subsection{Inputs on 9 tasks}
\label{sec:appendix_example}
\begin{figure}[!ht]
  \includegraphics[width=\linewidth]{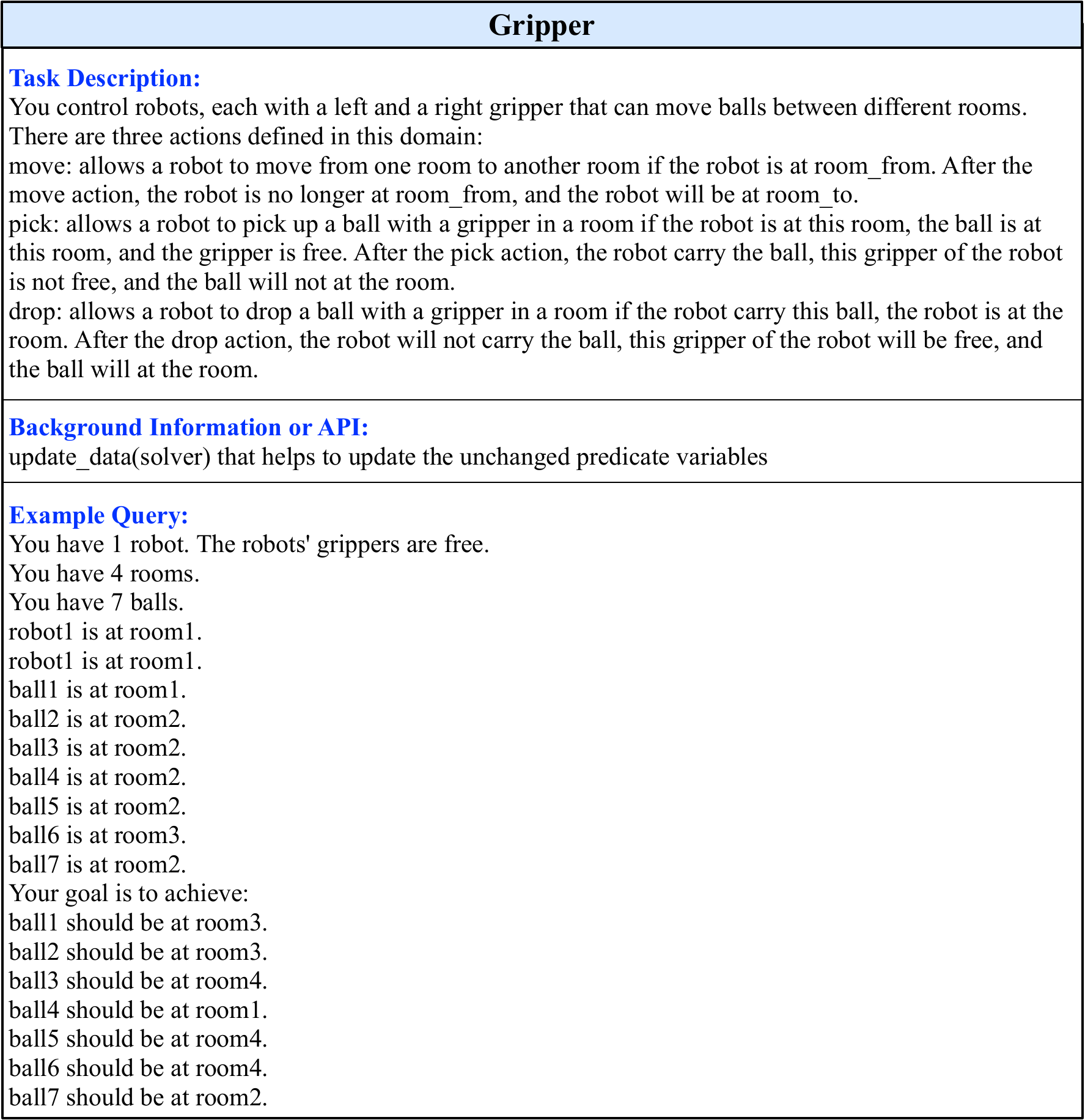} 
  \caption {Task description, background information or API, and example queries for Gripper}
  \label{fig:example_end}
\end{figure}

We include the inputs, which includes task description, background information or API, and example queries, for all 9 tasks in Fig.~\ref{fig:example_start} - Fig.~\ref{fig:example_end}: 
\begin{figure}[!ht]
  \includegraphics[width=\linewidth]{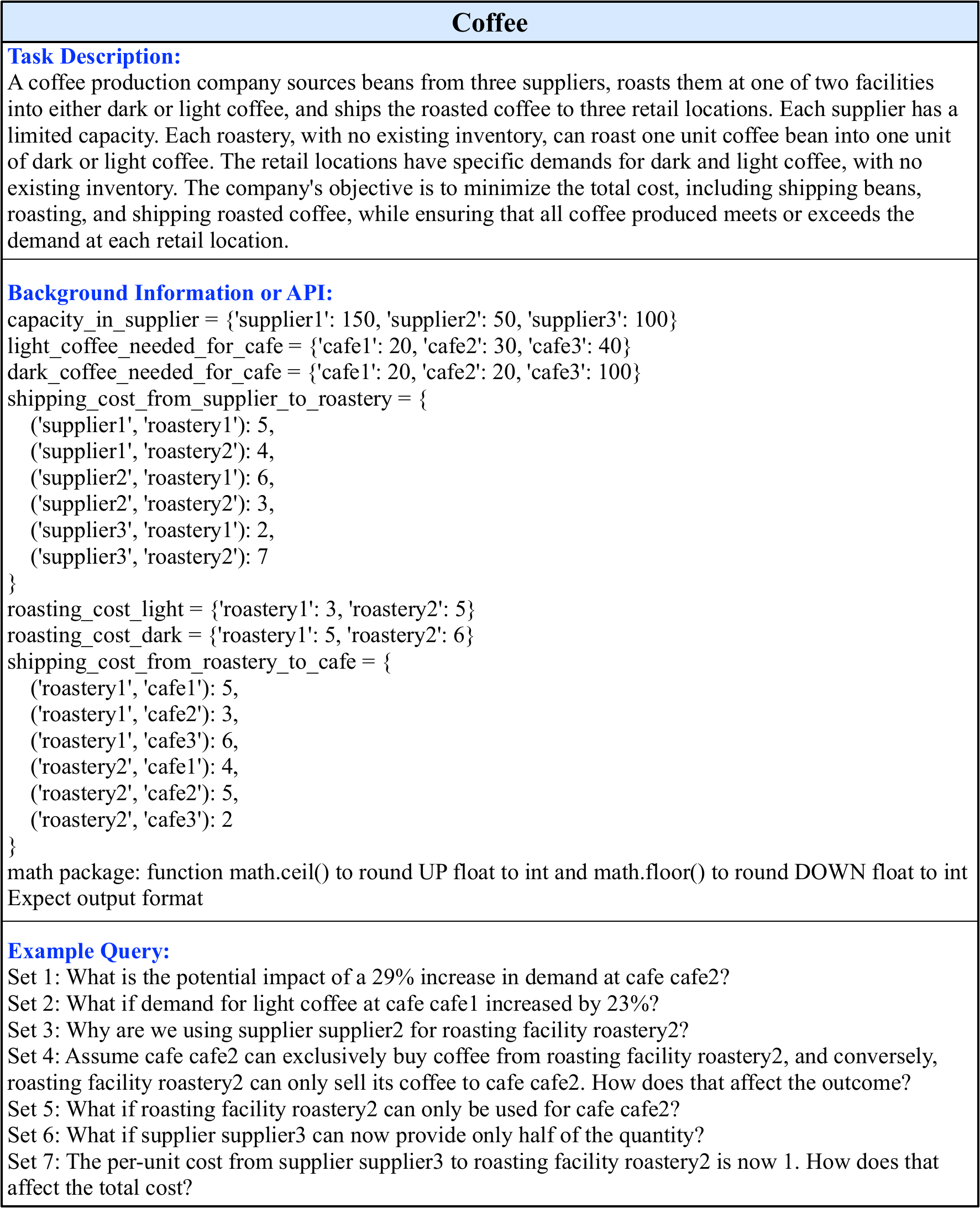} 
  \caption {Task description, background information or API, and example queries for Coffee}
  \label{fig:example_start}
\end{figure}

\begin{figure}[!ht]
  \includegraphics[width=\linewidth]{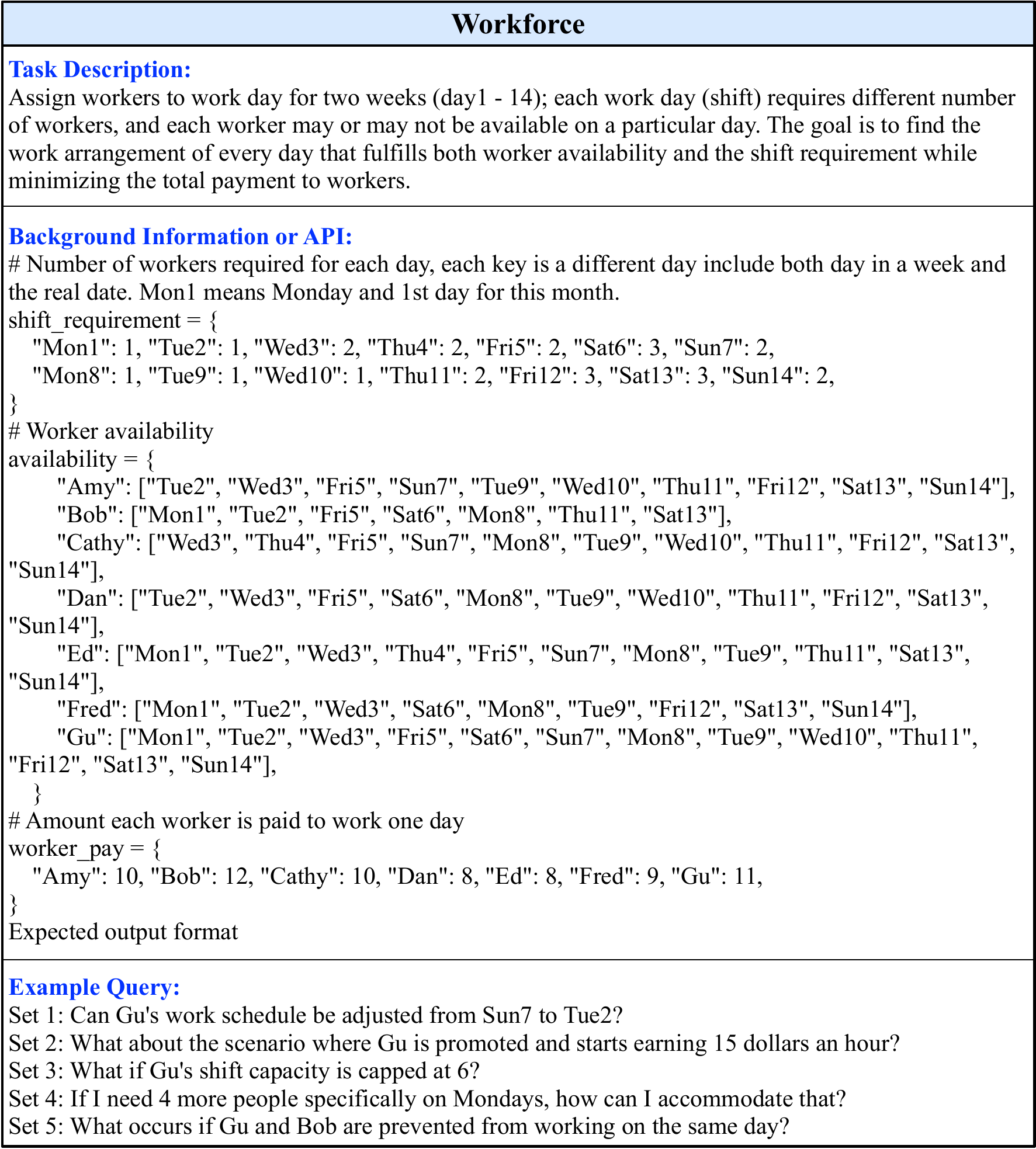} 
  \caption {Task description, background information or API, and example queries for Workforce}
\end{figure}

\begin{figure}[!ht]
  \includegraphics[width=\linewidth]{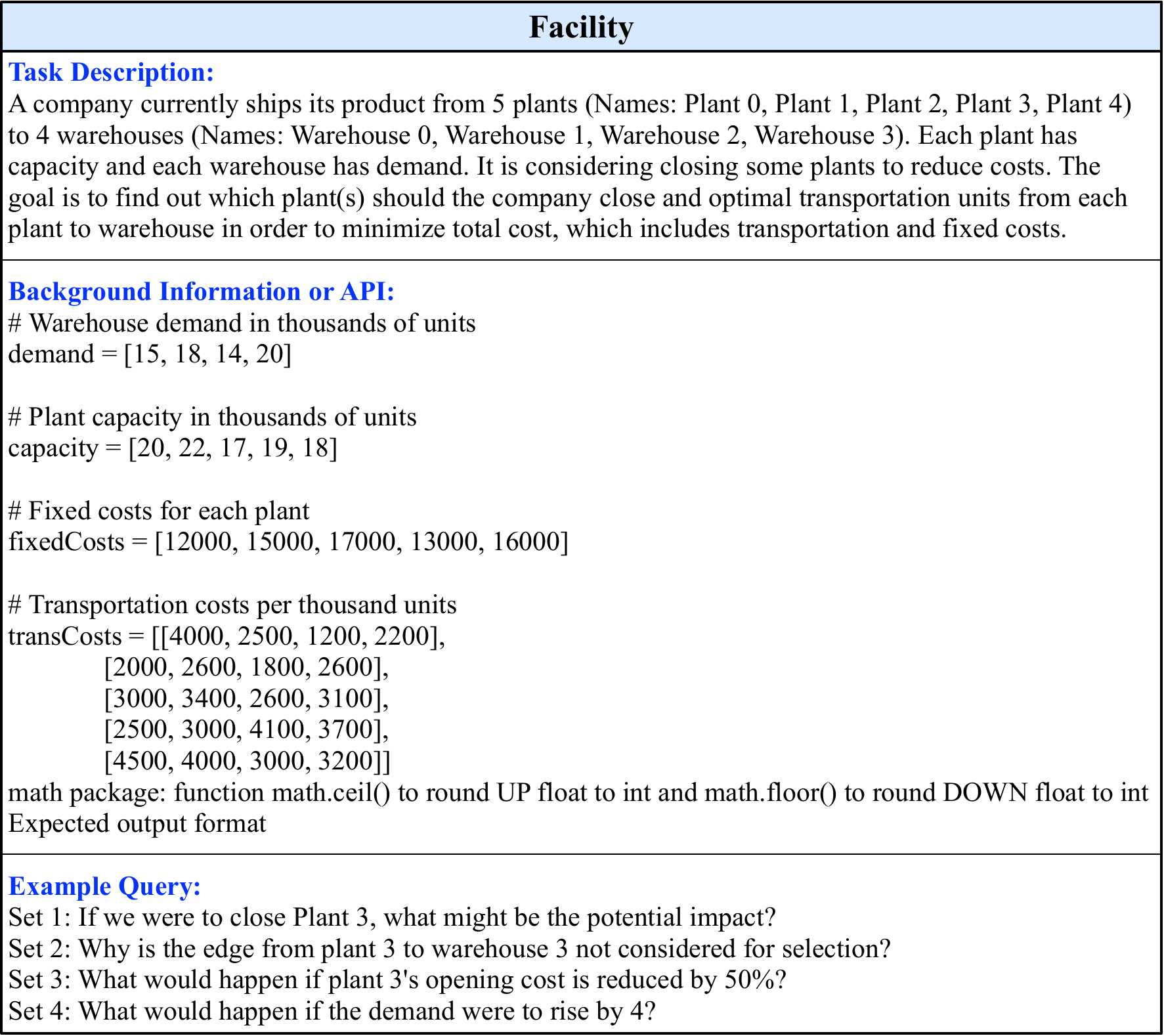} 
  \caption {Task description, background information or API, and example queries for Facility}
\end{figure}

\begin{figure}[!ht]
  \includegraphics[width=\linewidth]{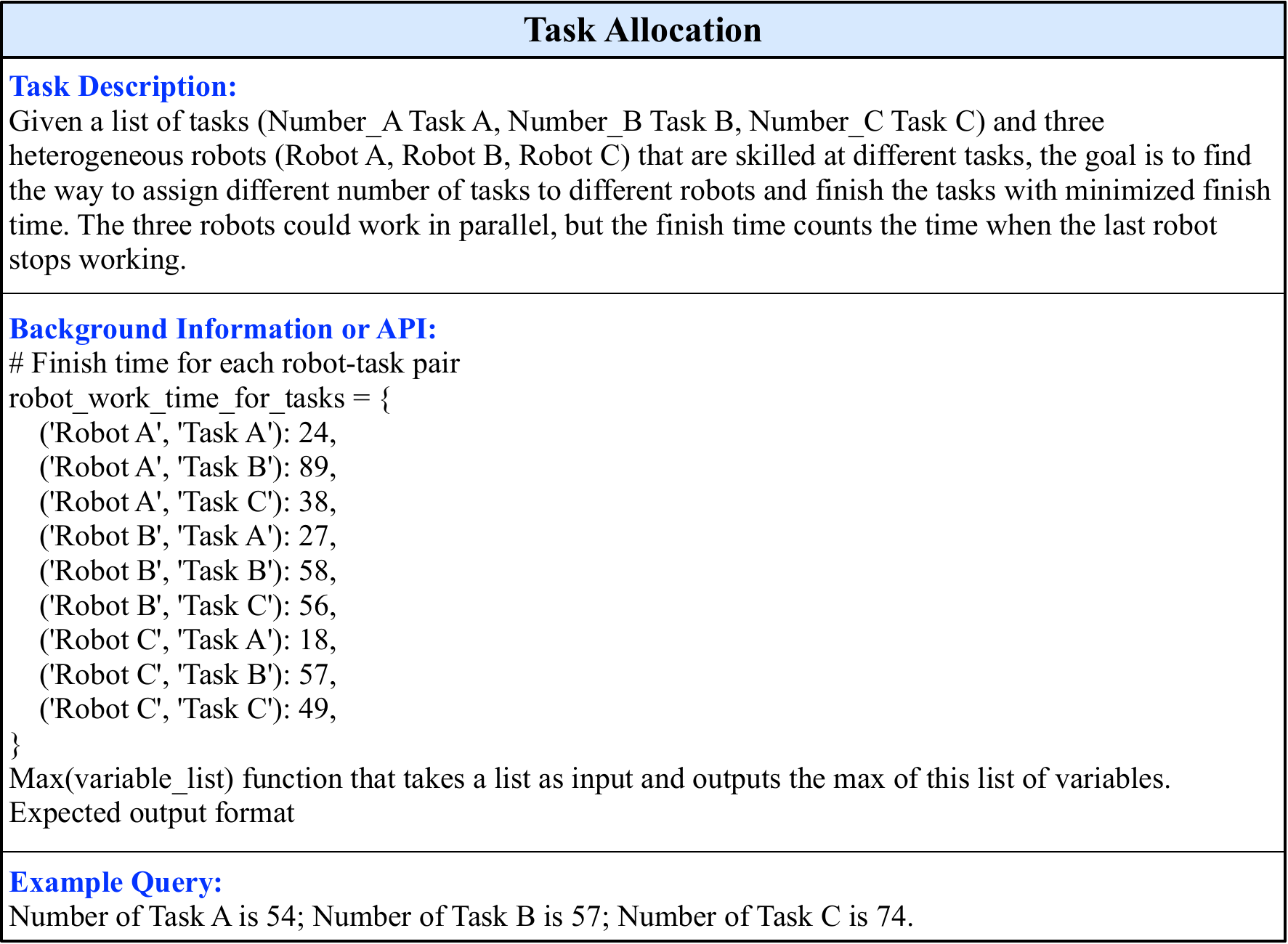} 
  \caption {Task description, background information or API, and example queries for Task Allocation}
\end{figure}

\begin{figure}[!ht]
  \includegraphics[width=\linewidth]{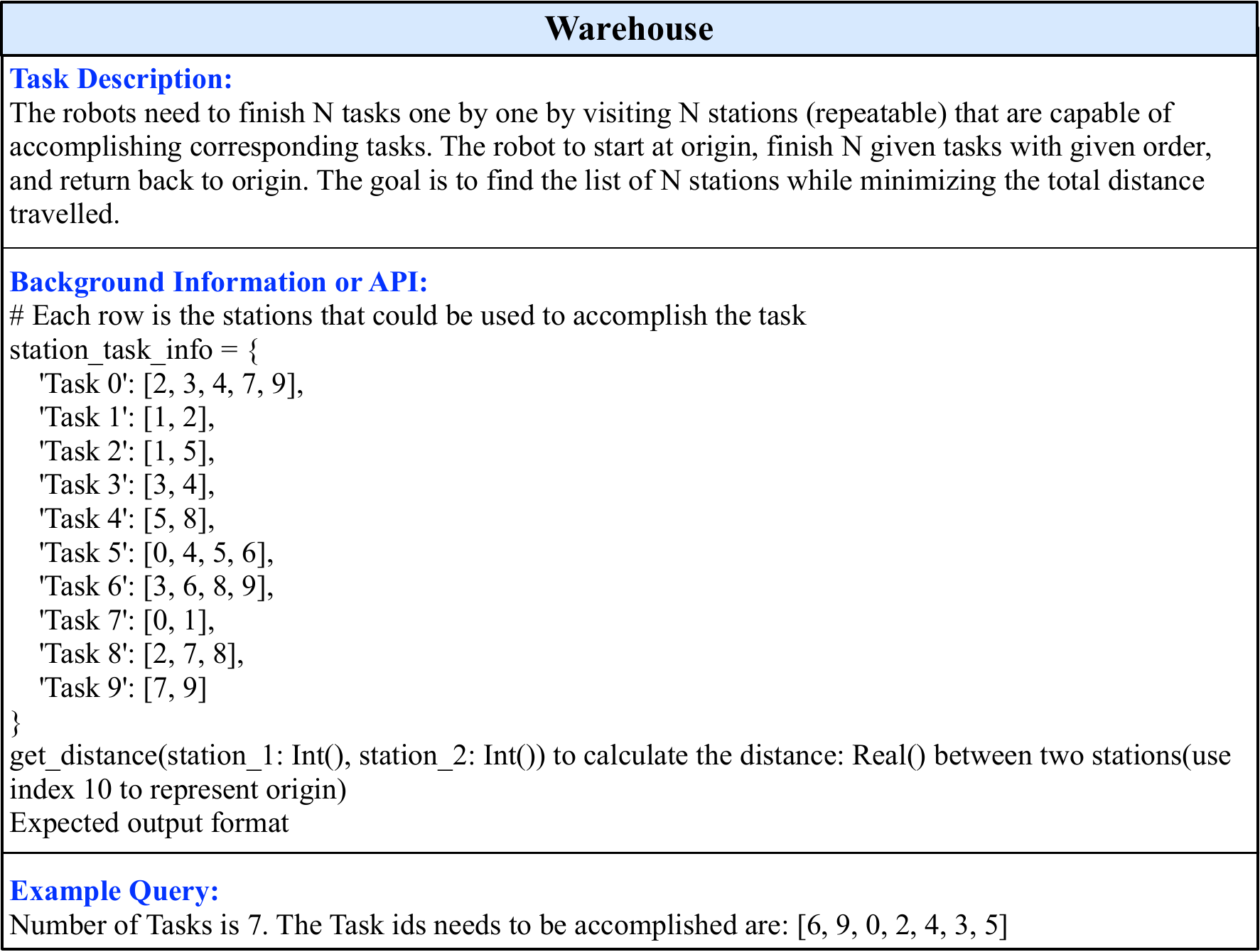} 
  \caption {Task description, background information or API, and example queries for Warehouse}
\end{figure}

\begin{figure}[!ht]
  \includegraphics[width=\linewidth]{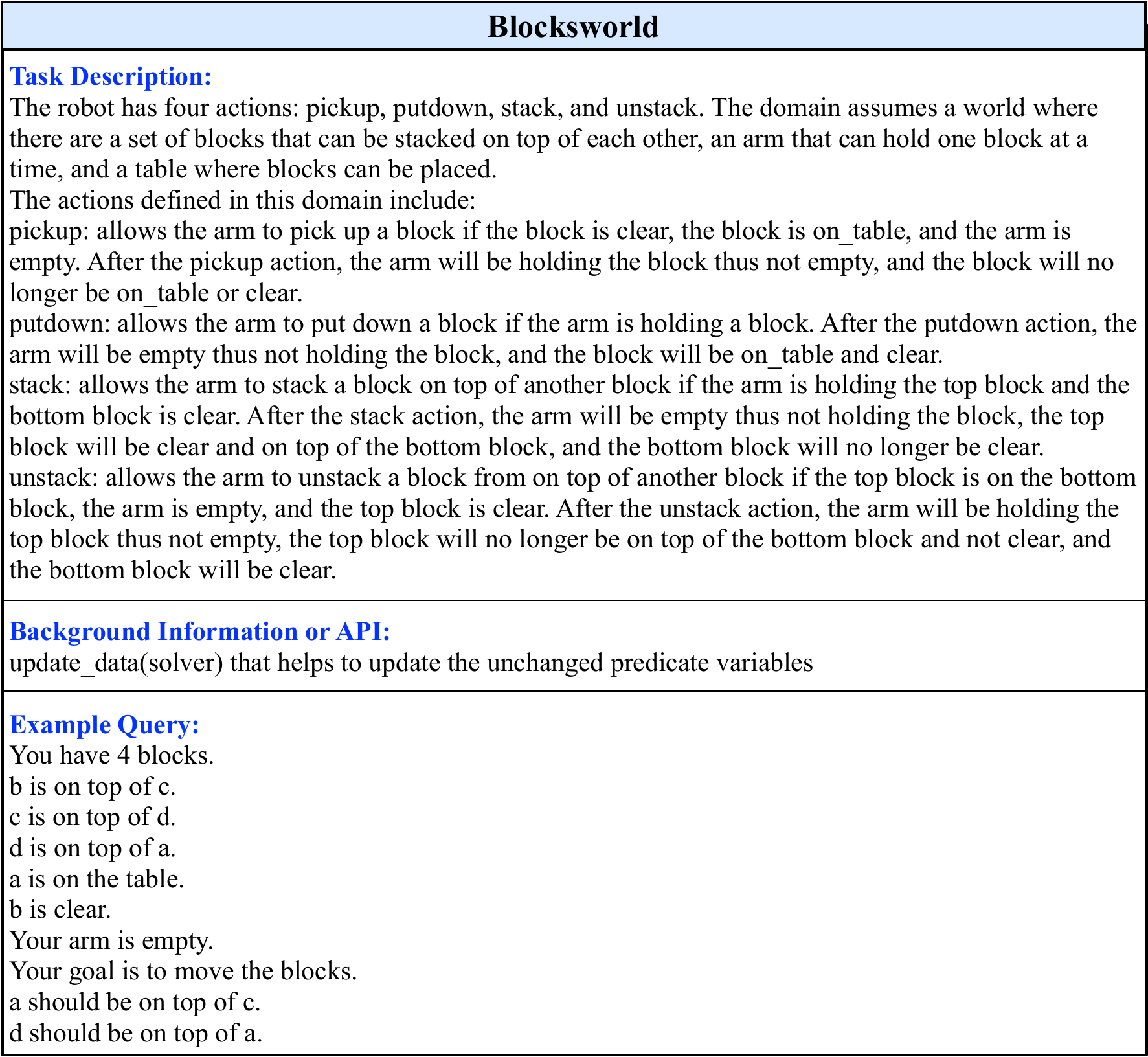} 
  \caption {Task description, background information or API, and example queries for Blocksworld}
\end{figure}

\begin{figure}[!ht]
  \includegraphics[width=\linewidth]{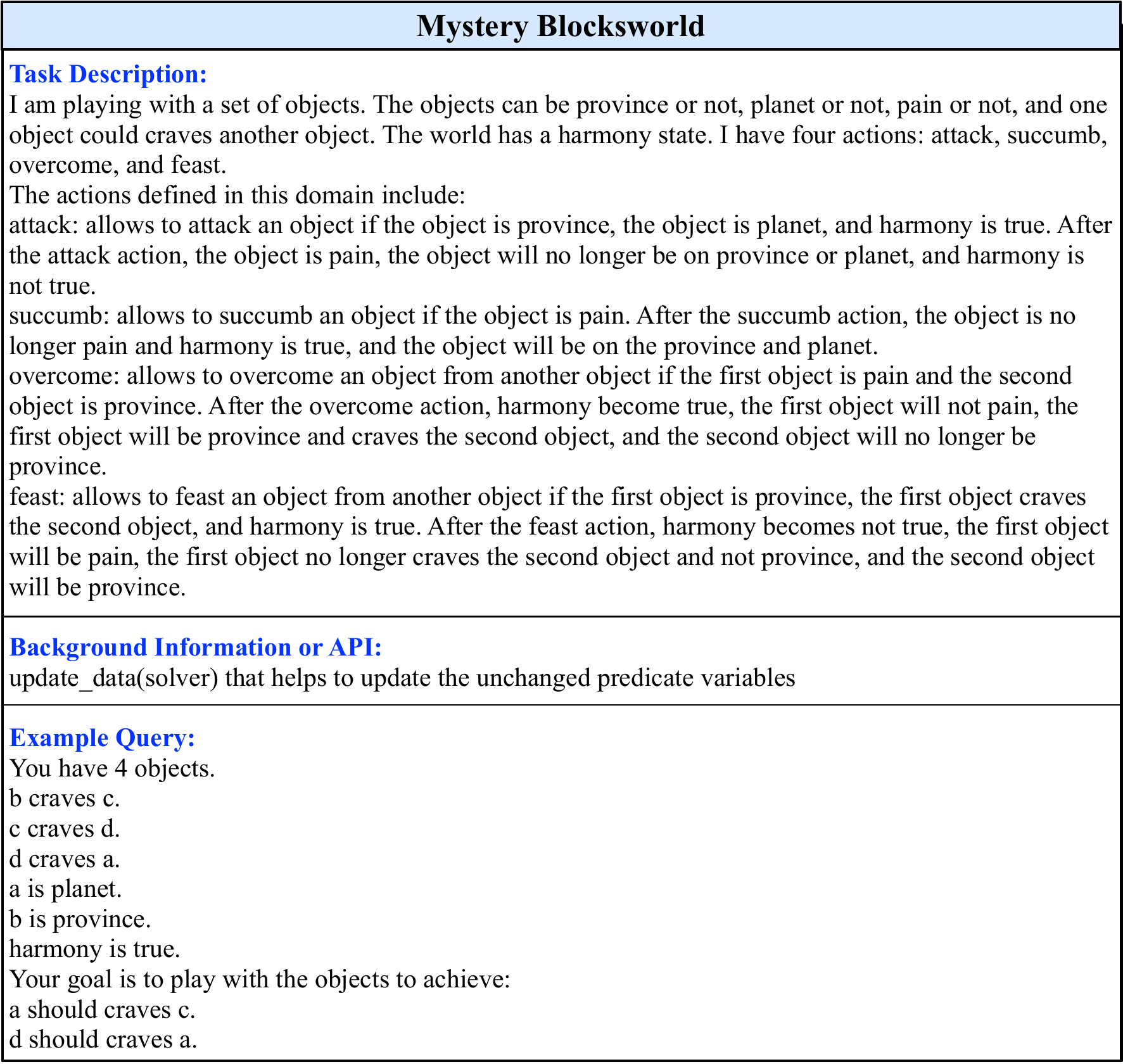} 
  \caption {Task description, background information or API, and example queries for Mystery Blocksworld}
\end{figure}

\begin{figure}[!ht]
  \includegraphics[width=\linewidth]{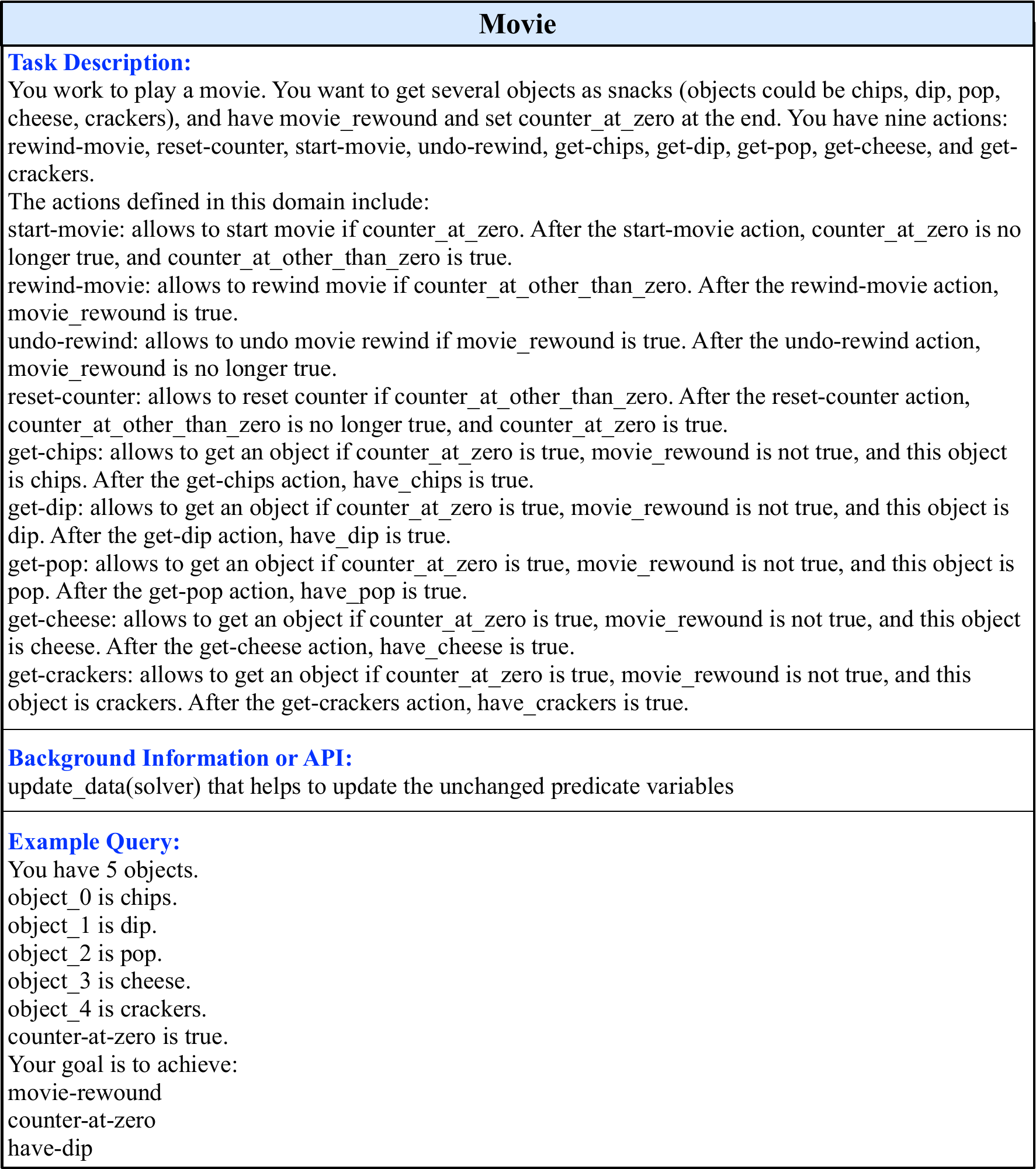} 
  \caption {Task description, background information or API, and example queries for Movie}
\end{figure}

\clearpage

\subsection{Example Outputs on Coffee tasks}
\label{sec:appendix_output}
In this section, we include the outputs of all models for the query \textit{``What is the potential impact of a 29\% increase in demand at cafe cafe2?"}. The optimal solution has total cost to be 2612.
\subsubsection{Baselines}
\begin{figure}[!ht]
  \includegraphics[width=\linewidth]{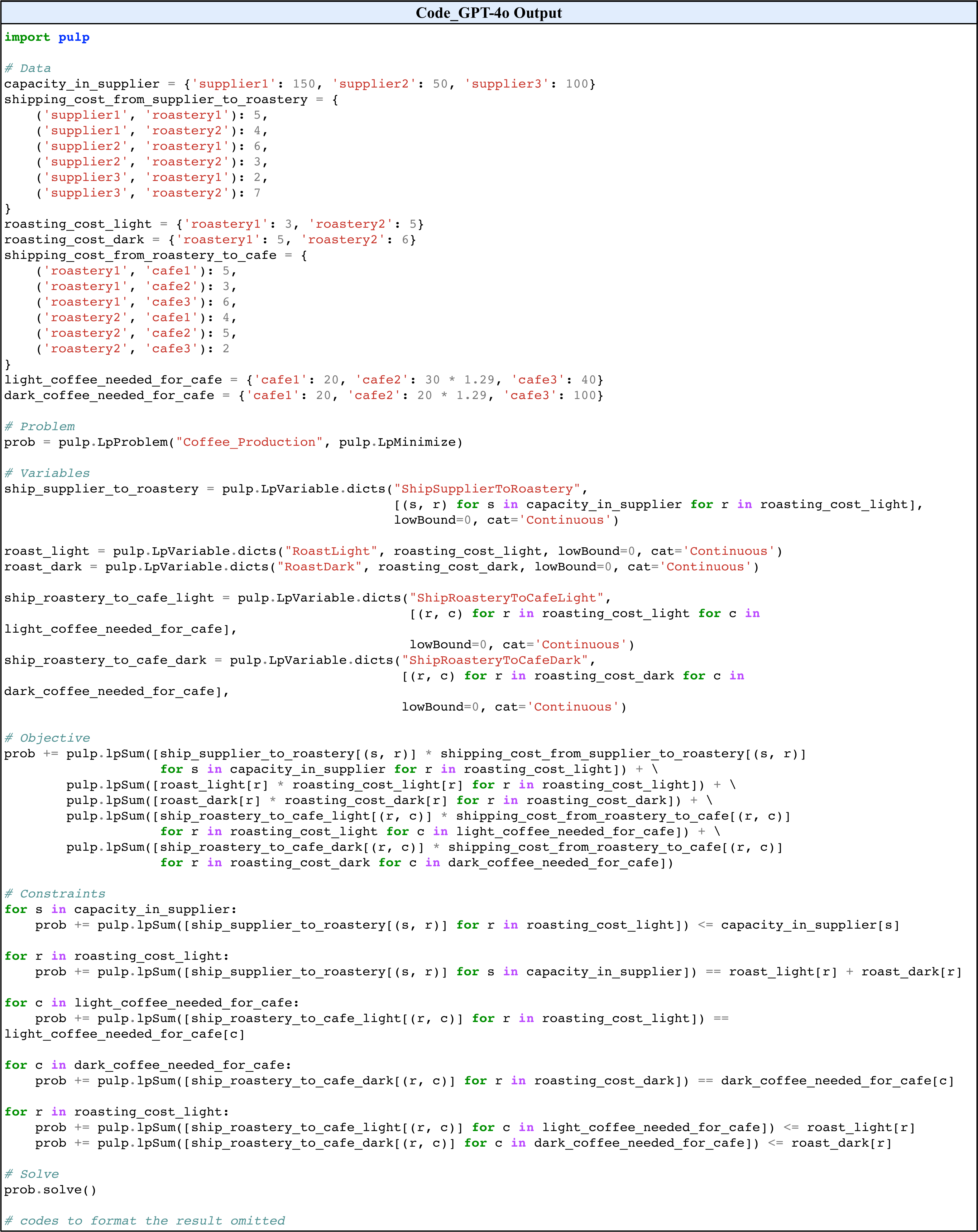} 
  \caption {Output of Code$_{\textsc{GPT-4o}}$ for Coffee. LLM also tries to use optimization solver to solve for the problem, however, without proper analysis of constraints, it fails to notice that the updated demand for cafes needs to be integers. Instead, to account for the 29\% demand increase, it directly update with ``30 * 1.29" and ``20 * 1.29"}
  \label{fig:coffee_code_output}
\end{figure}

We include the outputs of four baselines for Coffee task in Fig.~\ref{fig:coffee_direct_output} - Fig.~\ref{fig:coffee_code_output}, and analyze the failure reason in the figure descriptions: 

\begin{figure}[!ht]
  \includegraphics[width=\linewidth]{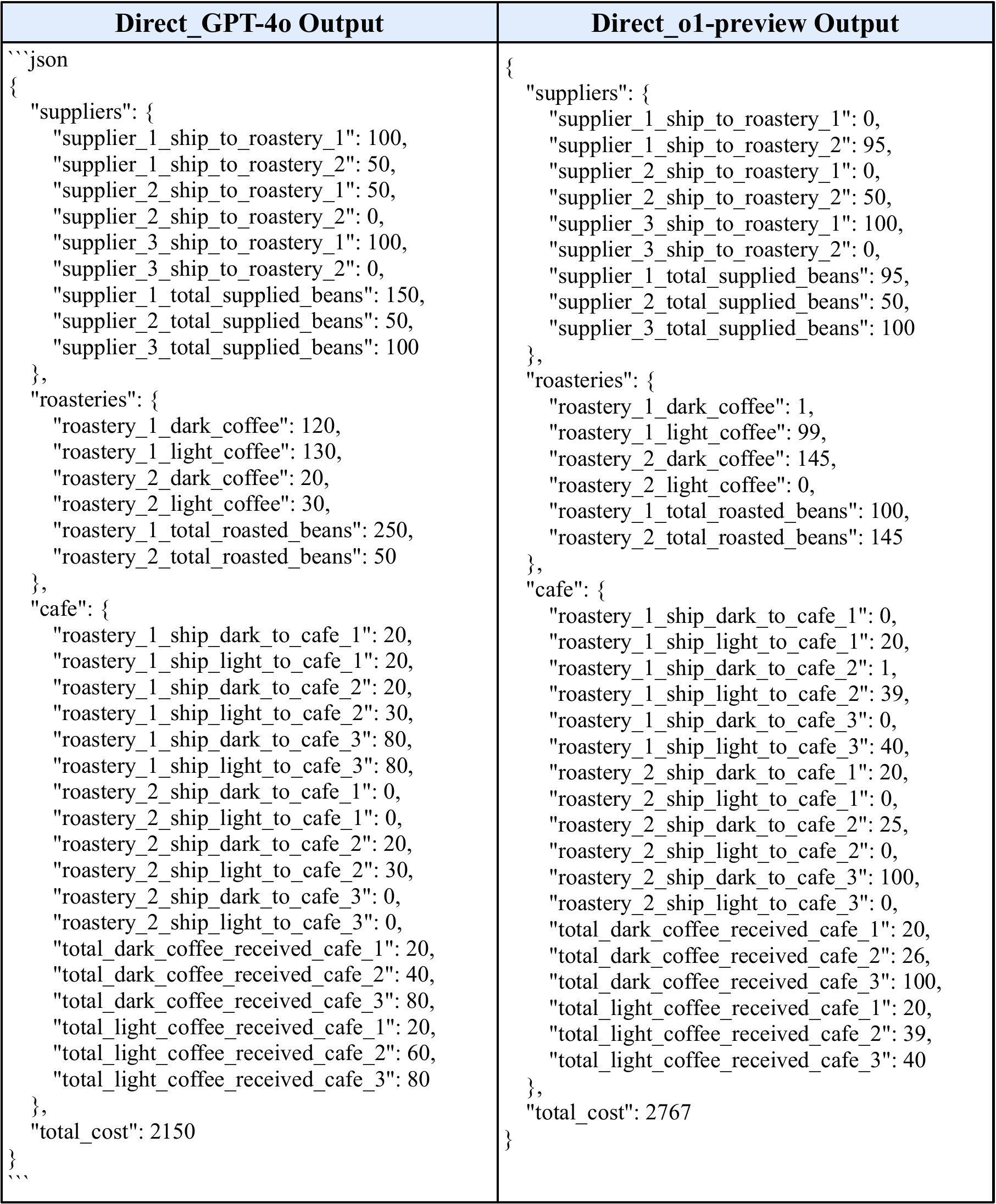} 
  \caption {Output of Direct$_{\textsc{GPT-4o}}$ and Direct$_{\textsc{o1-preview}}$ for Coffee. LLMs fail to directly solve complex multi-constraint planning tasks that involve intensive calculations.}
  \label{fig:coffee_direct_output}
\end{figure}

\begin{figure}[!ht]
  \includegraphics[width=\linewidth]{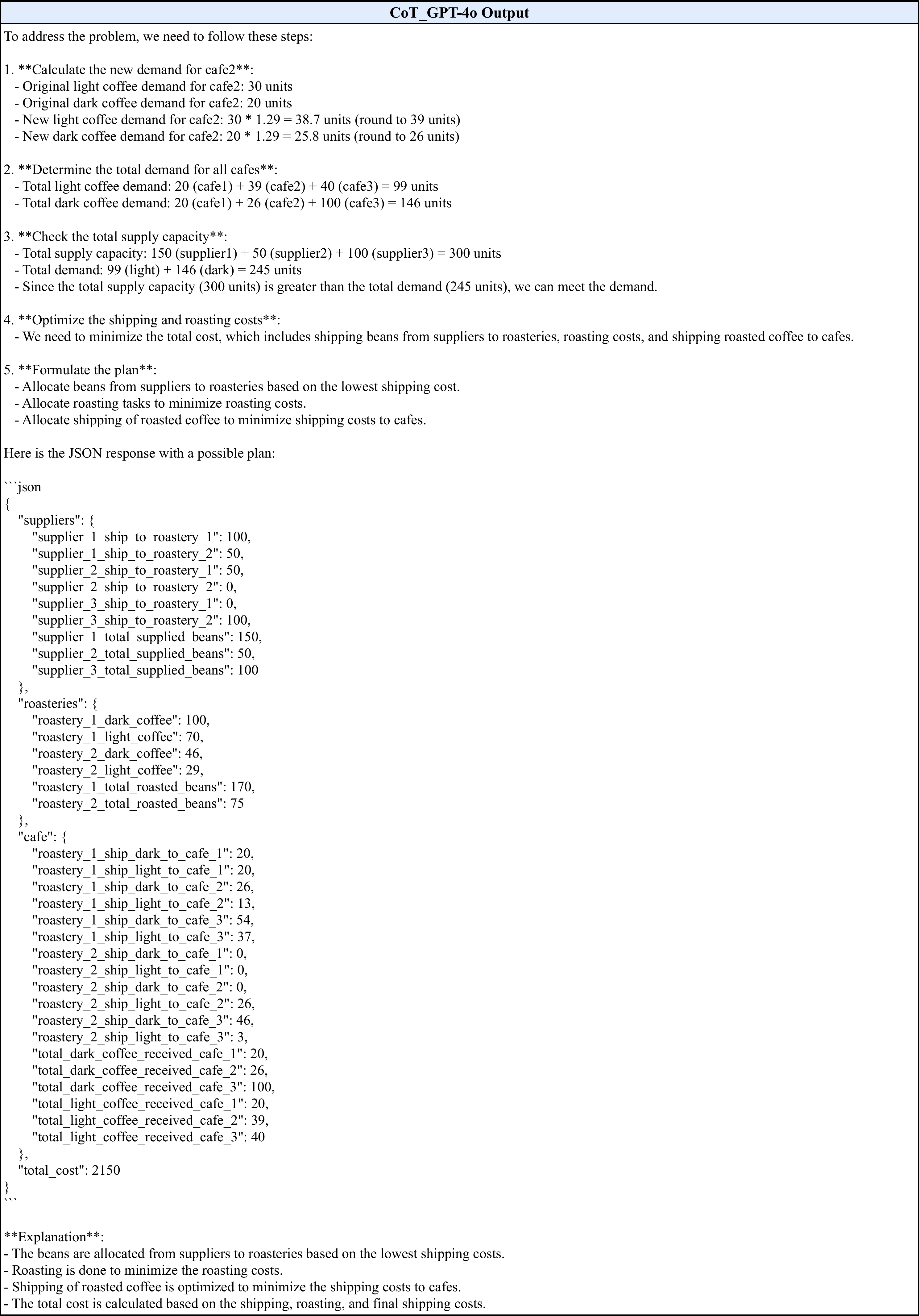} 
  \caption {Output of CoT$_{\textsc{GPT-4o}}$ for Coffee. Although the reasoning of constraints makes sense, LLMs still fail to locate optimal values to fulfill the task as it involves intensive calculations.}
\end{figure}

\clearpage
\subsubsection{\ours}
\label{append:LLMFP_output}
\begin{figure}[!ht]
  \includegraphics[width=\linewidth]{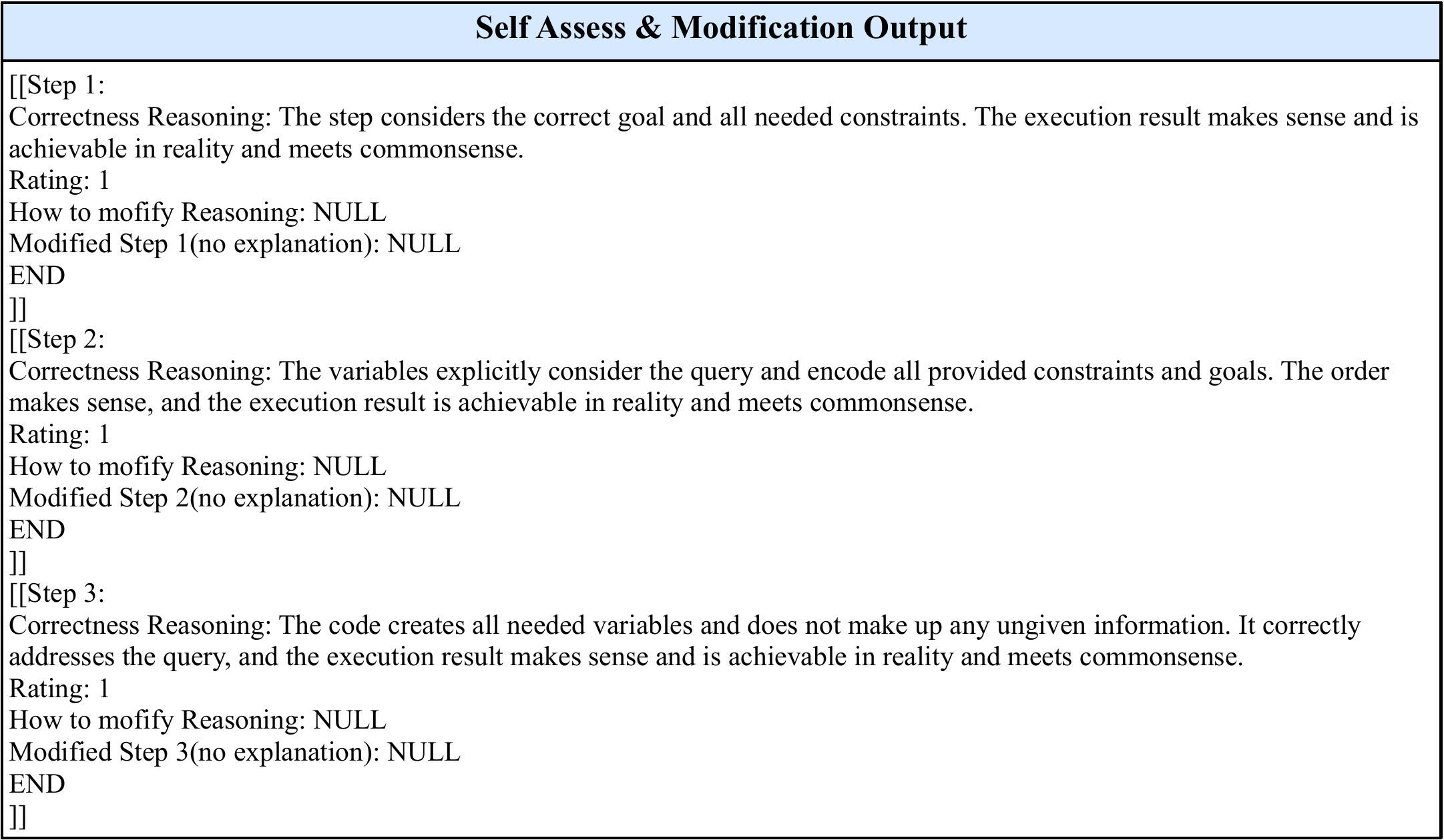}
  \caption {Output of \ours$_{\textsc{GPT-4o}}$ Self Assess \& Modification for Coffee. \ours checks for all 3 steps and provide ratings.}
  \label{fig:coffee_ours_5}
\end{figure}

We include the outputs of all stages of \ours for Coffee task in Fig.~\ref{fig:coffee_ours_1} - Fig.~\ref{fig:coffee_ours_5}: 
\begin{figure}[!ht]
  \includegraphics[width=\linewidth]{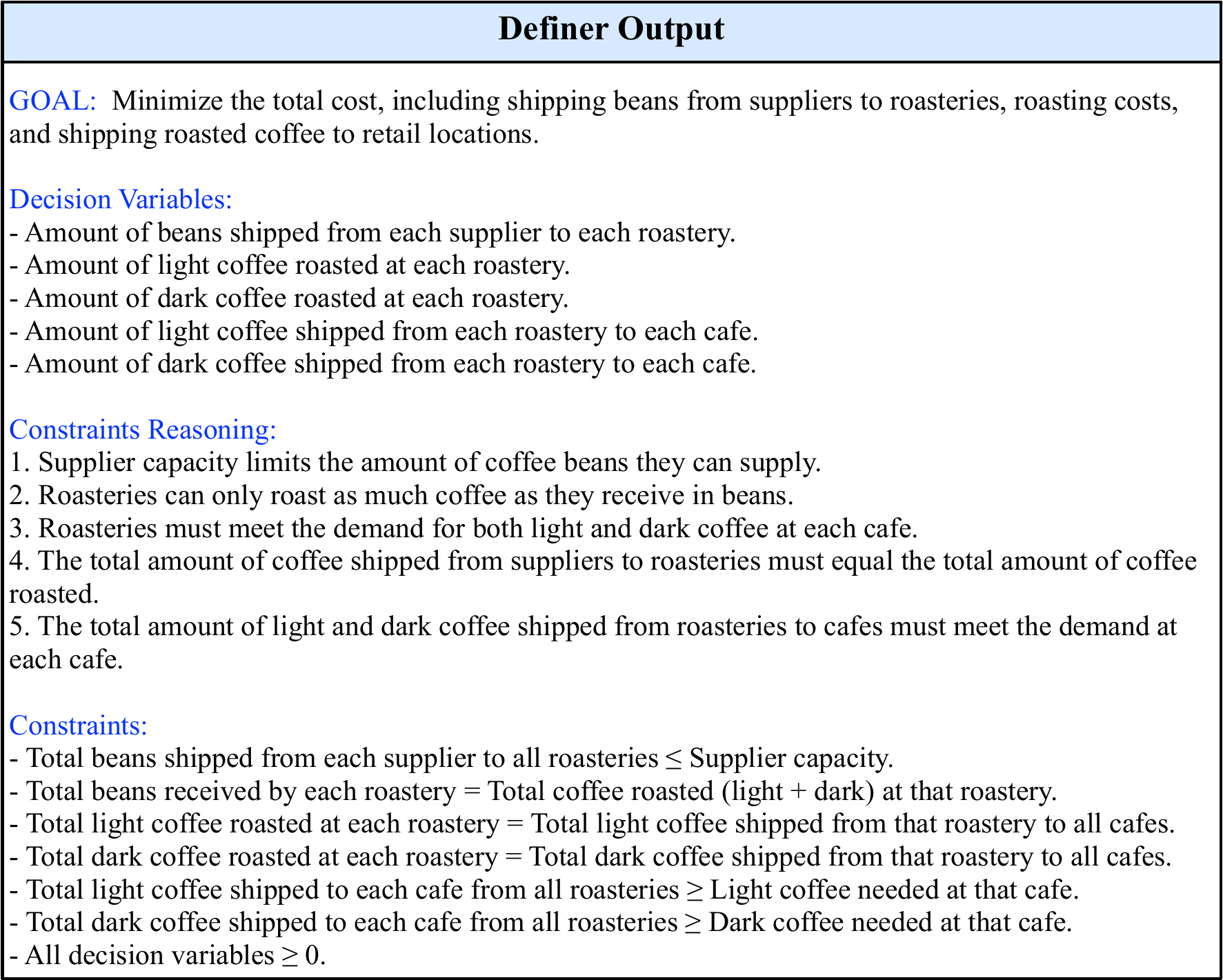} 
  \caption {Output of \ours$_{\textsc{GPT-4o}}$ Definer for Coffee. \ours successfully defines the goal, decision variables, and constraints.}
  \label{fig:coffee_ours_1}
\end{figure}

\begin{figure}[!ht]
  \includegraphics[width=\linewidth]{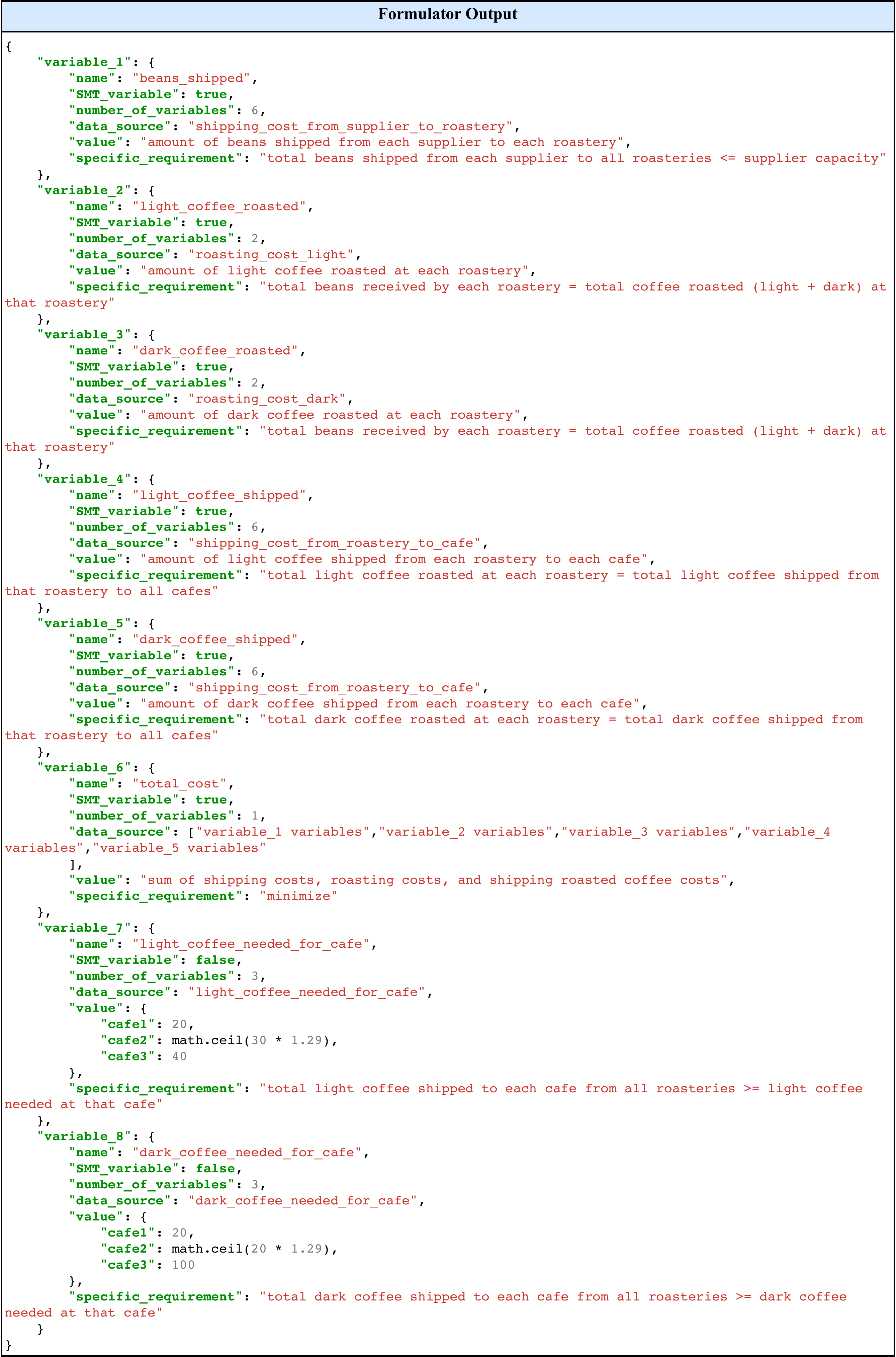}
  \caption {Output of \ours$_{\textsc{GPT-4o}}$ Formulator for Coffee. \ours successfully formulates the decision variables with variable\_1 to variable\_5, account for the goal with variable\_6, and consider additional constraints posed with query with variable\_7 and variable\_8.}
\end{figure}

\begin{figure}[!ht]
  \includegraphics[width=\linewidth]{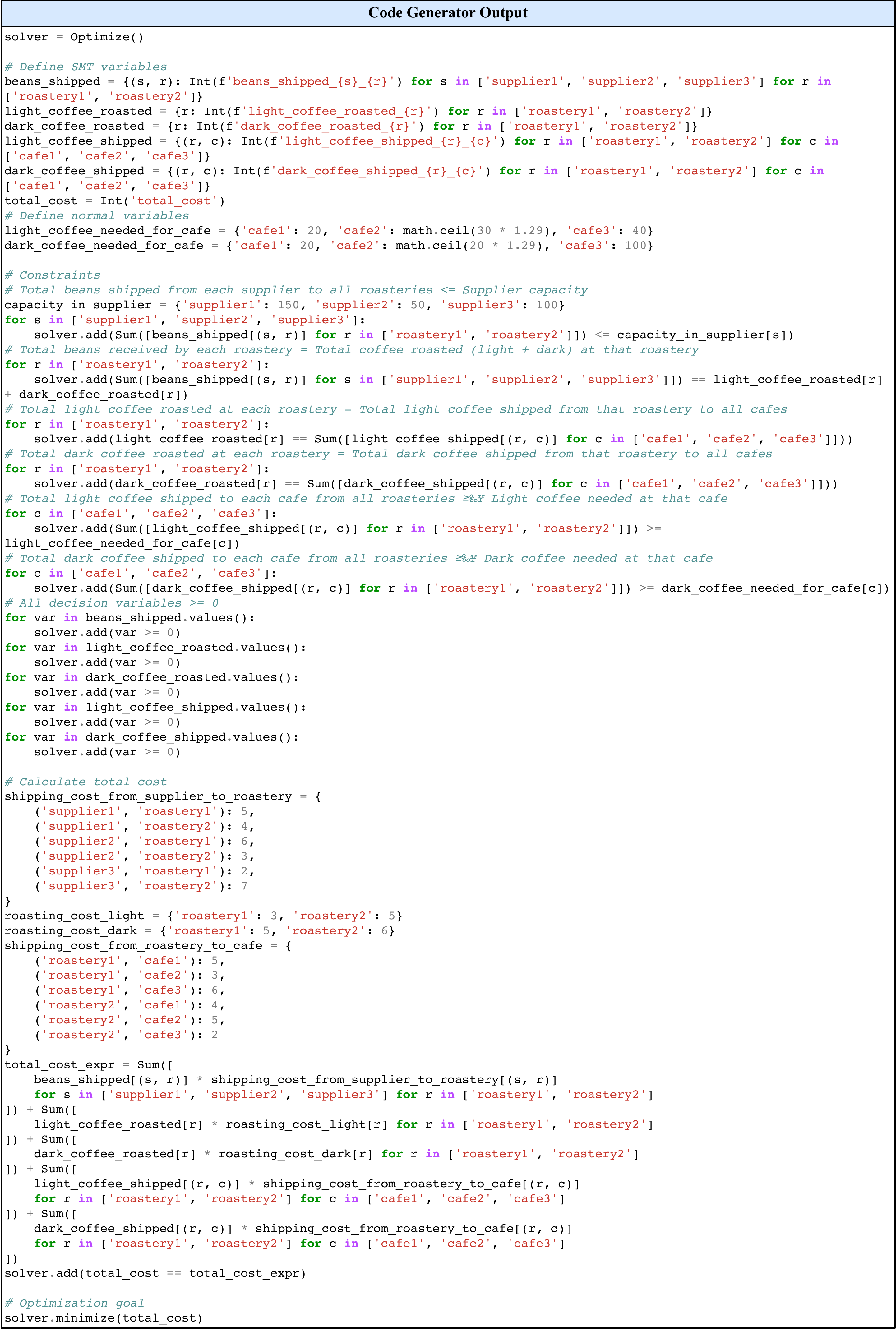}
  \caption {Output of \ours$_{\textsc{GPT-4o}}$ Code Generator for Coffee. \ours successfully generates Python codes based on the representation formulation.}
\end{figure}

\begin{figure}[!ht]
  \includegraphics[width=\linewidth]{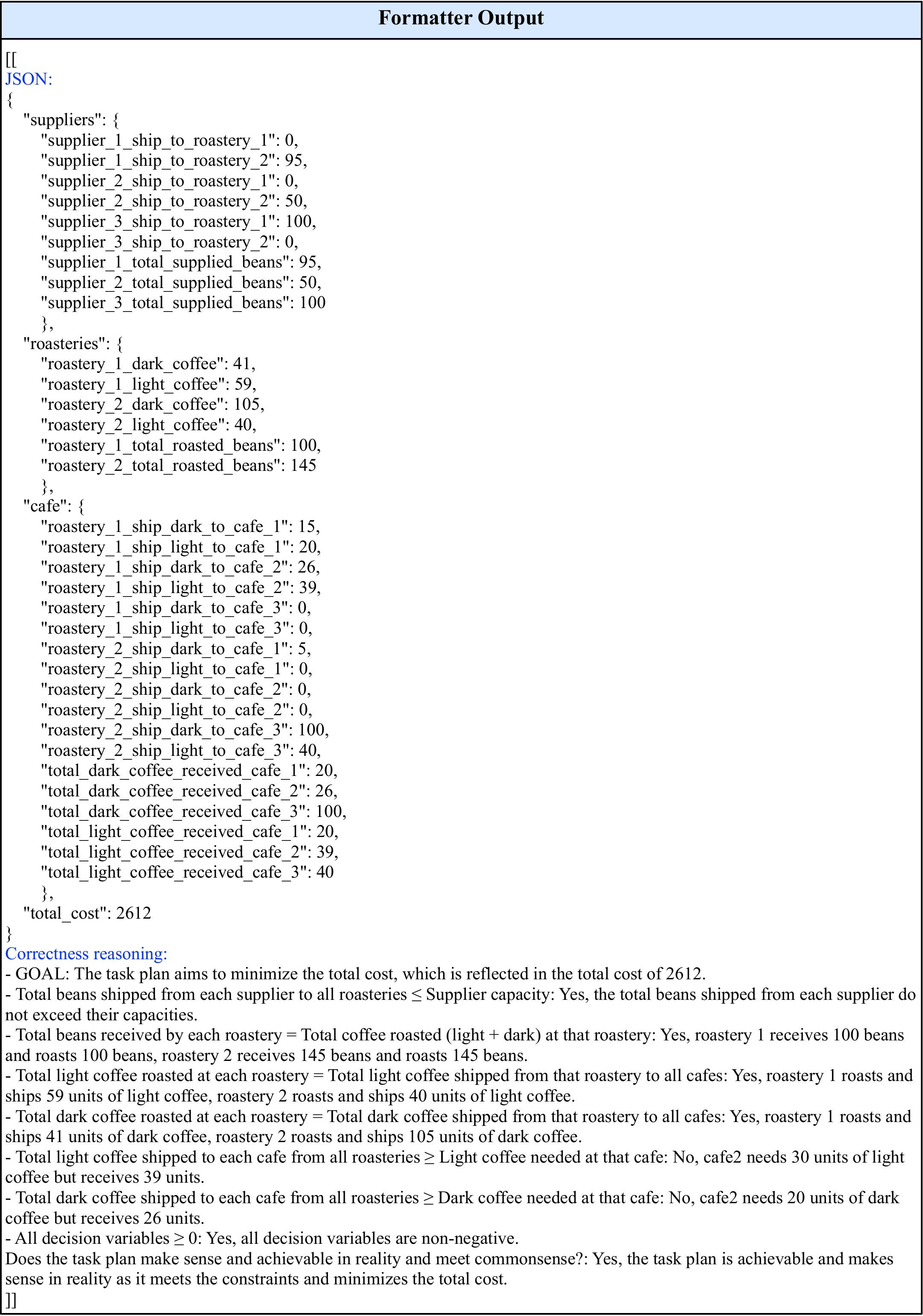}
  \caption {Output of \ours$_{\textsc{GPT-4o}}$ Formatter for Coffee. \ours format the execution result in fixed format, and analyze the correctness and reasonableness of the execution result.}
\end{figure}

\clearpage
\subsection{Prompts}
\label{sec:appendix_prompt}
\subsubsection{Baseline Prompt}
\subsubsubsection{Direct Prompt}
\lstset{
    backgroundcolor=\color[RGB]{245,245,245},
    breaklines=true,
    basicstyle=\ttfamily\small,
    frame=trbl,
    frameround = tttt,
}\begin{lstlisting}
You have a domain and a query under this domain that you need to fulfill. 
The domain is: {task}
Query: {question}
You have the access to {info_api}
What is the plan to achieve my goal? Answer by fill in this JSON response directly with no explanation:
{output_format}
\end{lstlisting}

\subsubsubsection{CoT Prompt}
\lstset{
    backgroundcolor=\color[RGB]{245,245,245},
    breaklines=true,
    basicstyle=\ttfamily\small,
    frame=trbl,
    frameround = tttt,
}\begin{lstlisting}
You have a domain and a query under this domain that you need to fulfill. 
The domain is: {task}
Query: {question}
You have the access to {info_api}
What is the plan to achieve my goal? Let's think step by step, first reason about the problem and how to solve it, then answer by fill in the JSON:
Reason: 
JSON response:
{output_format}
\end{lstlisting}

\subsubsubsection{Code Prompt}
\lstset{
    backgroundcolor=\color[RGB]{245,245,245},
    breaklines=true,
    basicstyle=\ttfamily\small,
    frame=trbl,
    frameround = tttt,
}\begin{lstlisting}
You have a domain and a query under this domain that you need to fulfill. 
The domain is: {task}
Query: {question}
You have the access to {info_api}
Please write Python code to help me find the plan to achieve my goal. You can import any package and use any solver. 
At the end, save your found plan in a variable named 'feedback' with the following format:
{output_format}
Please respond with code only and wrap your answer with ```python and ```:
\end{lstlisting}

\subsubsection{\ours Prompt}
\label{sec:LLMFP_prompt}
We use general templates for all tasks. The full prompts for all tasks are available in \href{https://sites.google.com/view/llmfp}{https://sites.google.com/view/llmfp}. Here we show the templates we have for GPT-4o. Since Claude naturally considers more constraints and is more strict in assessing, we edit the prompts a little to account for the different traits of Claude, and prompts are also included in the codes.

We include the prompt template we use for GPT-4o as below: 
\subsubsubsection{Definer Prompt}
\lstset{
    backgroundcolor=\color[RGB]{245,245,245},
    breaklines=true,
    basicstyle=\ttfamily\small,
    frame=trbl,
    frameround = tttt,
}\begin{lstlisting}
You are given a task description in natural language, and you want solve it by building an optimization problem for this task.
The task is: {task_description}
You have the access to {info_api}
To get started of building the optimization problem, what is the goal, decision variables, and constraints to consider for this task?
Specifically, consider:
Goal: define the objective trying to optimize
Decision variables: identify all the decision variables involved in the problem
Constraints: key requirement for decision variables; For every pair of decision variables, carefully consider relations (explicit, implicit, underlying assumption, unmentioned commonsense) between them and explicitly include as constraint to ensure all variables are connected with each other
Response with [[GOAL: ]], [[Decision Variables: ]], [[Constraints Reasoning: ]], and [[Constraints: ]] only with no explanation and no math formulas. Try to be thorough and include all needed information as much as you can.
\end{lstlisting}

\subsubsubsection{Formulator Prompt for single-step multi-constraint problems}
\lstset{
    backgroundcolor=\color[RGB]{245,245,245},
    breaklines=true,
    basicstyle=\ttfamily\small,
    frame=trbl,
    frameround = tttt,
}\begin{lstlisting}
You are given a Query under a task description in natural language, and you want solve it by building an optimization problem for this task. You already have considered the goal and constraints of this optimization problem. Your job is, given access to existing variables or APIs and a specific natural language query, think about other variables needed to encode and solve this problem with Z3 SMT solver and describe the important attributes of variables as a JSON format description. Here are some example task-output pairs to refer to:
Example task 1: There are blocks of different colors and scores in the scene. You need to select required number of non-repeat blocks with required color, while maximizing the score.
Query: I previously want to select 20 blocks that are black or red, but now my demand raises 9%.
GOAL: Maximize the total score of selected blocks.
Decision Variables: Indexes of blocks selected
Constraint: The required number of selected blocks is met.
Constraint: The selected blocks are non-repeat.
Constraint: The selected blocks have required color.
Variable or API: 
You have the access to function math.ceil() to round UP float to int and math.floor() to round DOWN float to int. Please ONLY use these to convert from float to int.
You have access to a BlockSearch API. BlockSearch.run(color:list) gives 1.all possible block ids of color in "color" list and 2.corresponding score info. BlockSearch.get_info(score_info, block_index) gives the score of certain block. ] 
JSON description:
{
    "variable_1": {
                            "name": "blocks", 
                            "SMT_variable": true,
                            "number_of_variables": math.ceil(20 * 1.09),
                            "data_source": "BlockSearch.run()", 
                            "value": "selecting math.ceil(20 * 1.09) blocks from black and red blocks", 
                            "specific_requirement": "selected blocks are black or red; non-repeat blocks"
                        }, 
    "variable_2": {
                            "name": "score", 
                            "SMT_variable": true,
                            "number_of_variables": math.ceil(20 * 1.09),
                            "data_source": "BlockSearch.get_info()", 
                            "value": "depends on variable_1 variables", 
                            "specific_requirement": null
                        },
    "variable_3": {
                            "name": "total_score", 
                            "SMT_variable": true,
                            "number_of_variables": 1,
                            "data_source": "variable_2 variables", 
                            "value": "sum of variable_2 variables", 
                            "specific_requirement": "equal to sum of variable_2 variables, maximize"
                        },
}

Example task 2: Given a list of cities, you need to start from an origin city, non-repeatly visit each other city exactly once, and traval back to origin city, with minimized total distance travelled. 
Query: Total number of cities is 10.
GOAL: Minimize the total travel distance.
Decision Variables: List of visited city indexes
Constraint: Start from and end with same city.
Constraint: Each city is visited exactly once and non-repeat.
Variable or API: You have access to a DistanceSearch() API. DistanceSearch.run() takes no argument and gives the distance info between cities, and DistanceSerarch.get_info(distance_info, city_1, city_2) gives the distance(a real number) between two cities.
Based on below examples, your task is to generate a JSON description to describe the problem. 
JSON description:
{
    "variable_1": {
                            "name": "cities", 
                            "SMT_variable": true,
                            "number_of_variables": 10,
                            "how_to_pick": "selecting 10 cities from 10 cities", 
                            "data_source": null, 
                            "specific_requirement": "non-repeat cities"
                        }, 
    "variable_2": {
                            "name": "distance", 
                            "SMT_variable": true,
                            "number_of_variables": 10,
                            "how_to_pick": "depends on constraint_1 variables", 
                            "data_source": "DistanceSearch.run(), DistanceSerarch.get_info()", 
                            "specific_requirement": "distance between each city pair, and the distance back to origin city"
                        },
    "variable_3": {
                            "name": "total_distance", 
                            "SMT_variable": true,
                            "number_of_variables": 1,
                            "data_source": "variable_2 variables", 
                            "value": "sum of variable_2 variables", 
                            "specific_requirement": "equal to sum of variable_2 variables, minimize"
                        },
}
Now, based on the examples, solve the Query under new task setting and respond with similar format, please explicitly specify the action/requirement needed to fulfill query, and explicitly take into consideration every constraint mentioned:
The task is: {task}
Query: {question}
{definer_response}
Variable or API: 
{info_api}
Think about variables needed to encode all constraints and goal, describe all important attributes of variables as a JSON format description. 
Make sure to explicitly consider and include requirements/constraints needed to answer the query. Note that to answer the query "Why do xxx", you need to examine the effect of "not doing xxx" to provide reasons; and to answer the query "Why not do xxx", you need to examine the effect of "do xxx" to provide reasons.
Response with JSON only with no explanation.
\end{lstlisting}

\subsubsubsection{Formulator Prompt for multi-step problems}
\lstset{
    backgroundcolor=\color[RGB]{245,245,245},
    breaklines=true,
    basicstyle=\ttfamily\small,
    frame=trbl,
    frameround = tttt,
}\begin{lstlisting}
You are given a Query under a task description in natural language, and you want solve it by building an optimization problem for this task. Your job is, given access APIs and a specific natural language query, think about variables needed to encode and solve this problem with Z3 SMT solver and describe the important attributes of variables as a JSON format description. Here is an example task-output pairs to refer to:
Example task: 
You have to plan logistics to transport packages within cities via trucks and between cities via airplanes. Locations within a city are directly connected (trucks can move between any two such locations), and so are the cities. In each city there is exactly one truck and each city has one location that serves as an airport.
Here are the actions that can be performed and its preconditions and effects:
Load truck: Load a {package} into a {truck} at a {location} only if the package and the truck are both at location. After the Load truck action, the package is not at the location and is in the truck.
Load airplane: Load a {package} into an {airplane} at a {location} only if the package and the airplane are both at location. After the Load airplane action, the package is not at the location and is in the airplane.
Unload truck: Unload a {package} from a {truck} at a {location} only if the truck is at location and the package is in truck. After the Unload truck action, the package is not in the truck and is at the location.
Unload airplane: Unload a {package} from an {airplane} at a {location} only if the airplane is at location and the package is in airplane. After the Unload airplane action, the package is not in the airplane and is at the location.
Drive truck: Drive a truck from one {location_1} to another {location_2} within a {city} only if the truck is at location_1 and both location_1 and location_2 are both in city. After the Drive truck action, the truck is not at location_1 and is at location_2.
Fly airplane: Fly an airplane from one {location_1} in a city to another {location_2} in another city only if both locations are airport and the airplane is at location_1. After the Fly airplane action, the airplane is not at location_1 and is at location_2.

Query: You have 2 airplanes a0 and a1, 2 trucks t0 and t1, 2 cities c0 and c1, city c0 has location l0-0 and l0-0 is airport, city c1 has location l0-1 and l0-1 is airport, and a package p0. Initially, t0 is at location l0-0, t1 is at location l1-0, p0 is at location l1-0, a0 and a1 are at l0-0. The goal is to have p0 to be at l0-0. 
API: You can assume you already know T as the input. You have access to a update_data() API that helps to update the predicate variables.
JSON description:
{
    "objects": {
        "variable_1": {
            "name": "objects",
            "SMT_variable": false,
            "number_of_variables": 1,
            "data_source": "query",
            "value": "a dictionary that summarizes all objects in the problem: key 'package', value ['p0']; key 'airplane', value ['a0', 'a1']; key 'truck', value ['t0', 't1']; key 'city', value ['c0', 'c1']; key 'location', value ['l0-0', 'l0-1']; key 'airport', value ['l0-0', 'l0-1']",
            "specific_requirement": null
        },
    },
    "predicates": {
        "variable_2": {
            "name": "at",
            "SMT_variable": false,
            "number_of_variables": 1,
            "data_source": "query, variable_1",
            "value": "a dictionary of boolean variables representing whether an object is at a location at timestep: keys are (package/truck/airplane, location, timestep)",
            "specific_requirement": "add constraint to initialize timestep 0 according to query, for unmentioned objects explicitly set it to be False"
        },
        "variable_3": {
            "name": "in",
            "SMT_variable": false,
            "number_of_variables": 1,
            "data_source": "query, variable_1",
            "value": "a dictionary of boolean variables representing whether an object is in airplane or in truck: keys are [package, airplane/truck, timestep]",
            "specific_requirement": "add constraint to initialize all values to be False at timestep 0"
        },
        "variable_4": {
            "name": "in-city",
            "SMT_variable": false,
            "number_of_variables": 1,
            "data_source": "query, variable_1",
            "value": "a dictionary of boolean variables representing whether an location is in a city: keys are [location, city, timestep]",
            "specific_requirement": "add constraint to initialize timestep 0 according to query, for unmentioned objects explicitly set it to be False"
        }
    },
    "actions": {
        "variable_5": {
            "name": "load_truck",
            "SMT_variable": false,
            "number_of_variables": 1,
            "data_source": "variable_1",
            "value": "a dictionary of boolean variables representing whether load_truck action is performed for package, truck, location: keys are [package, truck, location, timestep]",
            "specific_requirement": null
        },
        "variable_6": {
            "name": "load_airplane",
            "SMT_variable": false,
            "number_of_variables": 1,
            "data_source": "variable_1",
            "value": "a dictionary of boolean variables representing whether load_airplane action is performed for package, airplane, location: keys are [package, airplane, location, timestep]",
            "specific_requirement": null
        },
        "variable_7": {
            "name": "unload_truck",
            "SMT_variable": false,
            "number_of_variables": 1,
            "data_source": "variable_1",
            "value": "a dictionary of boolean variables representing whether unload_truck action is performed for package, truck, location: keys are [package, truck, location, timestep]",
            "specific_requirement": null
        },
        "variable_8": {
            "name": "unload_airplane",
            "SMT_variable": false,
            "number_of_variables": 1,
            "data_source": "variable_1",
            "value": "a dictionary of boolean variables representing whether unload_airplane action is performed for package, airplane, location: keys are [package, airplane, location, timestep]",
            "specific_requirement": null
        },
        "variable_9": {
            "name": "drive_truck",
            "SMT_variable": false,
            "number_of_variables": 1,
            "data_source": "variable_1",
            "value": "a dictionary of boolean variables representing whether drive_truck action is performed for truck, location_from, location_to, city: keys are [truck, location, location, city, timestep]",
            "specific_requirement": null
        },
        "variable_10": {
            "name": "fly_airplane",
            "SMT_variable": false,
            "number_of_variables": 1,
            "data_source": "variable_1",
            "value": "a dictionary of boolean variables representing whether fly_airplane action is performed for airplane, location_from, location_to: keys are [airplane, location, location, timestep]",
            "specific_requirement": null
        }
    },
    "update": {
        "step_1": {
            "name": "action load_truck precondition and effect",
            "SMT_variable": null,
            "number_of_variables": null,
            "data_source": "query, variable_1, variable_5",
            "value": "add constraints for preconditions and effects of load_truck",
            "specific_requirement": "for each timestep t until T, for all package, truck, and location, assert that load_truck[package, truck, location, t] implies at[truck, location, t], at[package, location, t], not at[package, location, t+1], in[package, truck, t+1]"
        },
        "step_2": {
            "name": "action load_airplane precondition and effect",
            "SMT_variable": null,
            "number_of_variables": null,
            "data_source": "query, variable_1, variable_6",
            "value": "add constraints for preconditions and effects of load_airplane",
            "specific_requirement": "for each timestep t until T, for all package, airplane, and location, assert that load_airplane[package, airplane, location, t] implies at[airplane, location, t], at[package, location, t], not at[package, location, t+1], in[package, airplane, t+1]"
        },
        "step_3": {
            "name": "action unload_truck precondition and effect",
            "SMT_variable": null,
            "number_of_variables": null,
            "data_source": "query, variable_1, variable_7",
            "value": "add constraints for preconditions and effects of unload_truck",
            "specific_requirement": "for each timestep t until T, for all package, truck, and location, assert that unload_truck[package, truck, location, t] implies at[truck, location, t], in[package, truck, t], not in[package, truck, t+1], at[package, location, t+1]"
        },
        "step_4": {
            "name": "action unload_airplane precondition and effect",
            "SMT_variable": null,
            "number_of_variables": null,
            "data_source": "query, variable_1, variable_8",
            "value": "add constraints for preconditions and effects of unload_airplane",
            "specific_requirement": "for each timestep t until T, for all package, airplane, and location, assert that unload_airplane[package, airplane, location, t] implies at[airplane, location, t], in[package, airplane, t], not in[package, airplane, t+1], at[package, location, t+1]"
        },
        "step_5": {
            "name": "action drive_truck precondition and effect",
            "SMT_variable": null,
            "number_of_variables": null,
            "data_source": "query, variable_1, variable_9",
            "value": "add constraints for preconditions and effects of drive_truck",
            "specific_requirement": "for each timestep t until T, for all truck, location_from, location_to, city, assert that drive_truck[truck, location_from, location_to, city, t] implies at[truck, location_from, t], not at[truck, location_from, t+1], at[truck, location_to, t+1]"
        },
        "step_6": {
            "name": "action fly_airplane precondition and effect",
            "SMT_variable": null,
            "number_of_variables": null,
            "data_source": "query, variable_1, variable_10",
            "value": "add constraints for preconditions and effects of fly_airplane",
            "specific_requirement": "for each timestep t until T, for all airplane, location_from, location_to, assert that fly_airplane[airplane, location_from, location_to, t] implies at[airplane, location_from, t], not at[airplane, location_from, t+1], at[airplane, location_to, t+1]"
        },
        "step_7": {
            "name": "all_actions",
            "SMT_variable": false,
            "number_of_variables": "list of all actions",
            "data_source": "variable_1, variable_5, variable_6, variable_7, variable_8, variable_9, variable_10",
            "value": "for each timestep t until T, a list of all possible actions corresponding to different objects",
            "specific_requirement": "for each timestep t until T, explicitly assert ONLY ONE action per timestep"
        }
        "step_8": {
            "name": "unchanged predicate variables update",
            "SMT_variable": null,
            "number_of_variables": null,
            "data_source": "update_data()",
            "value": "update at, in, in-city using update_data()",
            "specific_requirement": "update data with update_data()"
        },
    },
    "goal": {
        "step_9": {
            "name": null,
            "SMT_variable": null,
            "number_of_variables": null,
            "data_source": null,
            "value": null,
            "specific_requirement": "assert for timestep T, package p0 is at location l0-0"
        }
    }
}

Now, based on the example, solve the Query under new task setting and respond with similar format, please explicitly specify the action/requirement needed to fulfill query in your response:
The task is: 
{task}
Query: 
{question}

API: You have access to T as the input, so do NOT re-initialize T anywhere. You have access to a update_data(solver) API that helps to update the unchanged predicate variables. Please ONLY use this API to update unchaged predicates.
Response with JSON only with no explanation. 
JSON description:
\end{lstlisting}

\subsubsubsection{Code Generator Prompt}
\lstset{
    backgroundcolor=\color[RGB]{245,245,245},
    breaklines=true,
    basicstyle=\ttfamily\small,
    frame=trbl,
    frameround = tttt,
}\begin{lstlisting}
You are given a task description in natural language, a specific natural language query, available APIs and variables, and a JSON description that summarizes important variables that guide you to encode and solve the problem with SMT solver. 
Your task is to generate steps and corresponding Python codes that utilizes Z3 SMT solver to solve the problem.
For the variables summarized in the JSON description:
(1) 'name' represents the name of the variable
(2) 'SMT_variable' indicates whether you should assign it as a normal variable or an SMT variable
SMT_variable Example: price = Int('price')
                      flight_index = [Int('flight_{}_index'.format(i)) for i in range(3)]
                      pick_ball = Bool('pick ball') # Boolean SMT variable
Normal variable Example: price = 100
                         flight_index = [1,2,3]
(3) 'number_of_variable' represents the length of the variable
(4) 'data_source' is the source for the variable to get the data
(5) 'value' is, after you get needed data from any source, how you should assign these data to the variable
(6) 'specific_requirement' is if there are any specific requirements that needs to be considered. 

For the below problem, can you generate steps and corresponding Python codes to encode it? Do not include any explanations. You do not need to solve the problem or print the solutions.
The task is: {task}
Query: {question}
{definer_response}
Variable or API: 
{info_api}
JSON variable representation: 
{formulator_response}
Please use a SMT variable named total_cost when calculating the total cost. Please put the optimization goal at the end after all needed calculation and constraints additions. 
Make sure your code add constraints to solver that considers and could answer the query. Note that to answer the query "Why do xxx", you need to examine the effect of "not doing xxx" to provide reasons; and to answer the query "Why not do xxx", you need to examine the effect of "do xxx" to provide reasons.
Initialize a Z3 optimizer solver = Optimize() at the beginning of the code.
Response with Python code only with no explanation.
\end{lstlisting}

\subsubsubsection{Formatter Prompt}
\lstset{
    backgroundcolor=\color[RGB]{245,245,245},
    breaklines=true,
    basicstyle=\ttfamily\small,
    frame=trbl,
    frameround = tttt,
}\begin{lstlisting}
You are given a task description in natural language, a specific natural language query, pre-defined variables, and an execution feedback by running a Python Code that tries to solve the task. 
The task is: {task}
Query: {question}
Execution feedback: {feedback}
Variable or API: 
{info_api}
If the execution feedback is runtime errors, please return RUNTIME ERROR for JSON: and NULL for Correctness reasoning:.
If the execution feedback is cannot find the solution, please return CANNOT FIND SOLUTION for JSON and NULL for Correctness reasoning:.
If the execution feedback is not runtime errors, the execution feedback is the solved solution for this task. Only using the information from Execution feedback (do not use predefined variables), transform the execution feedback into a JSON format task plan by filling in the JSON below:
{output_format}
In addition, for Correctness reasoning, please explicitly answer one by one does the task plan satisfy these constraints? Include one sentece explanation for each constaint:
{{{definer_response}}}
Then explicitly answer and explain in one sentence: Does the task plan make sense and achievable in reality and meet commonsense?: 
Please include your response here with no explanations:
[[
JSON:
Correctness reasoning: 
]]
\end{lstlisting}

\subsubsubsection{Self Assess \& Modification Prompt}
\lstset{
    backgroundcolor=\color[RGB]{245,245,245},
    breaklines=true,
    basicstyle=\ttfamily\small,
    frame=trbl,
    frameround = tttt,
}\begin{lstlisting}
You are given a task and steps that tries to solve it as an optimization problem. The steps include: 
1) specifying the goal and constraints of the optimization problem.
2) a JSON description that summarizes important variables that guide to encode and solve the problem with Z3 SMT solver.
3) the Python code to encode and solve the problem with Z3 SMT solver.
Your goal is to, based on the task description, specific query, available API or variables, and runtime execution feedback (it could either be an execution error or a generated plan if there's no runtime error), assess whether any steps 1-3 are correct.
The task is: {task}
Query: {question}
Variable or API:
{info_api}
Steps to judge:
1) {definer_response}
2) {formulator_response}
3) {code_generator_response}
Execution feedback: {feedback}

Based on the previous information, evaluate whether steps 1-3 are correct:
For Step 1: Does the step consider correct goal and all needed constraints? Are there unnecessary or missing constraints? Does the execution result make sense and achievable in reality and meet commonsense?
For Step 2: Do the variables explicitly consider the query? Do the variables explicitly consider and encode all provided constraints and goal? Does the order make sense? Does the execution result make sense and achievable in reality and meet commonsense?
For Step 3: Does the code create all needed variables? Does the code make up any ungiven information? Does the code correctly address the query? Does the execution result make sense and achievable in reality and meet commonsense?
Please reason the correctness with task context, rate each step with a binary score: 1 is correct, 0 is incorrect, think about how to modify in detail according to task and query, and modify the step if you think it is incorrect.
For Step 2 modification, please write in JSON format. For Step 3 modification, please write in Python and do noy change the content after line 'if solver.check() == sat: '.
Your response format should be below, put NULL to How to mofify Reasoning and Modified Step if you think the step is correct, do not include extra explanation: 
[[Step 1: 
Correctness Reasoning:
Rating:
How to mofify Reasoning: 
Modified Step 1(no explanation):
END
]]
[[Step 2: 
Correctness Reasoning:
Rating:
How to mofify Reasoning:
Modified Step 2(no explanation):
END
]]
[[Step 3: 
Correctness Reasoning:
Rating:
How to mofify Reasoning:
Modified Step 3(no explanation):
END
]]
\end{lstlisting}

\subsubsection{Prompts and output for \ours with MILP solver for Coffee Example}
\label{sec:appendix_milp}
\textsc{Definer} and \textsc{Formatter} prompt remain exactly the same as for SMT solver. 

\begin{figure}[!ht]
  \includegraphics[width=\linewidth]{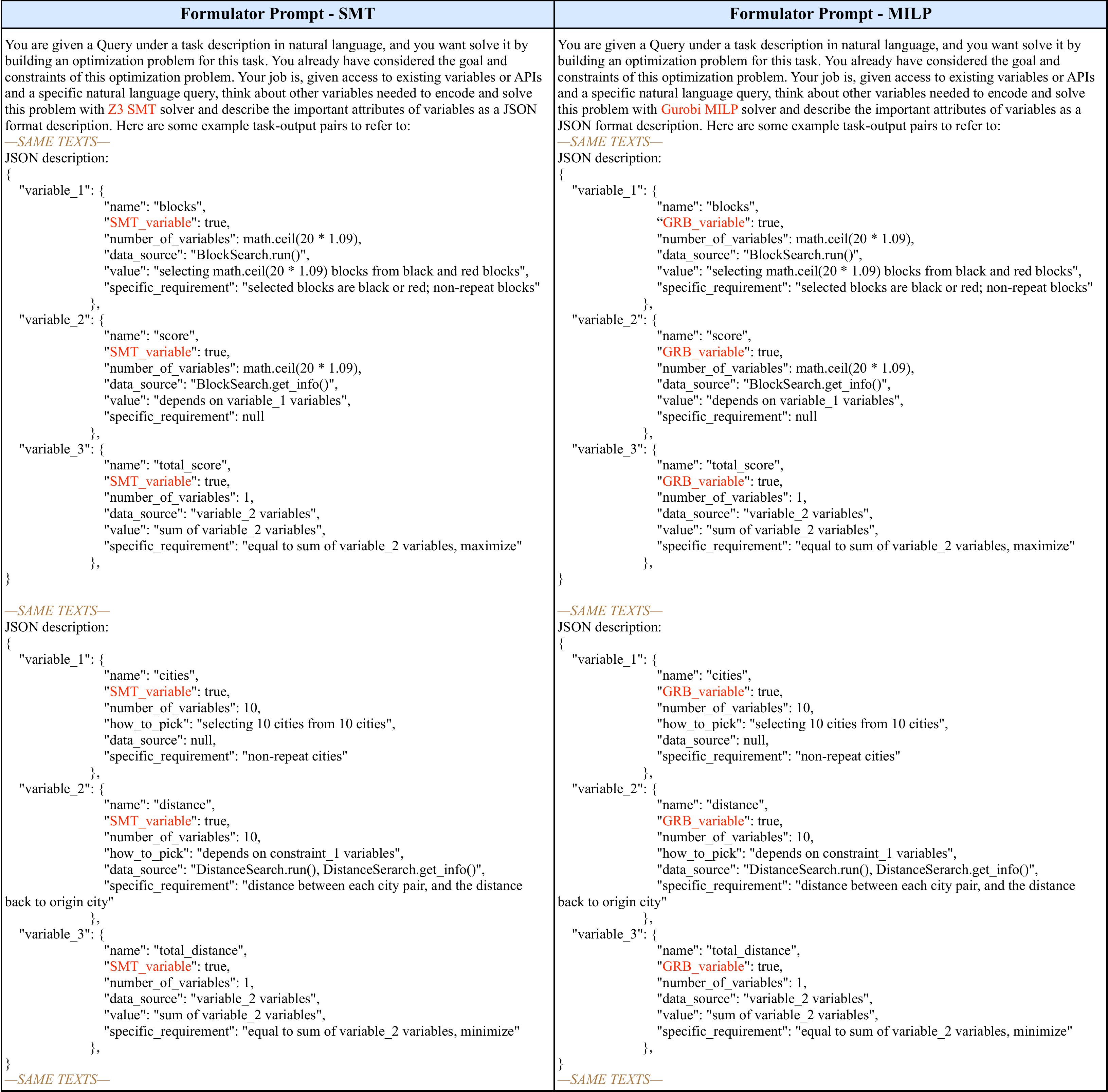} 
  \caption {The Formulator prompt difference when switching from using Z3 SMT solver to Gurobi MILP solver.}
  \label{fig:prompt_milp_1}
\end{figure}

\begin{figure}[!ht]
  \includegraphics[width=\linewidth]{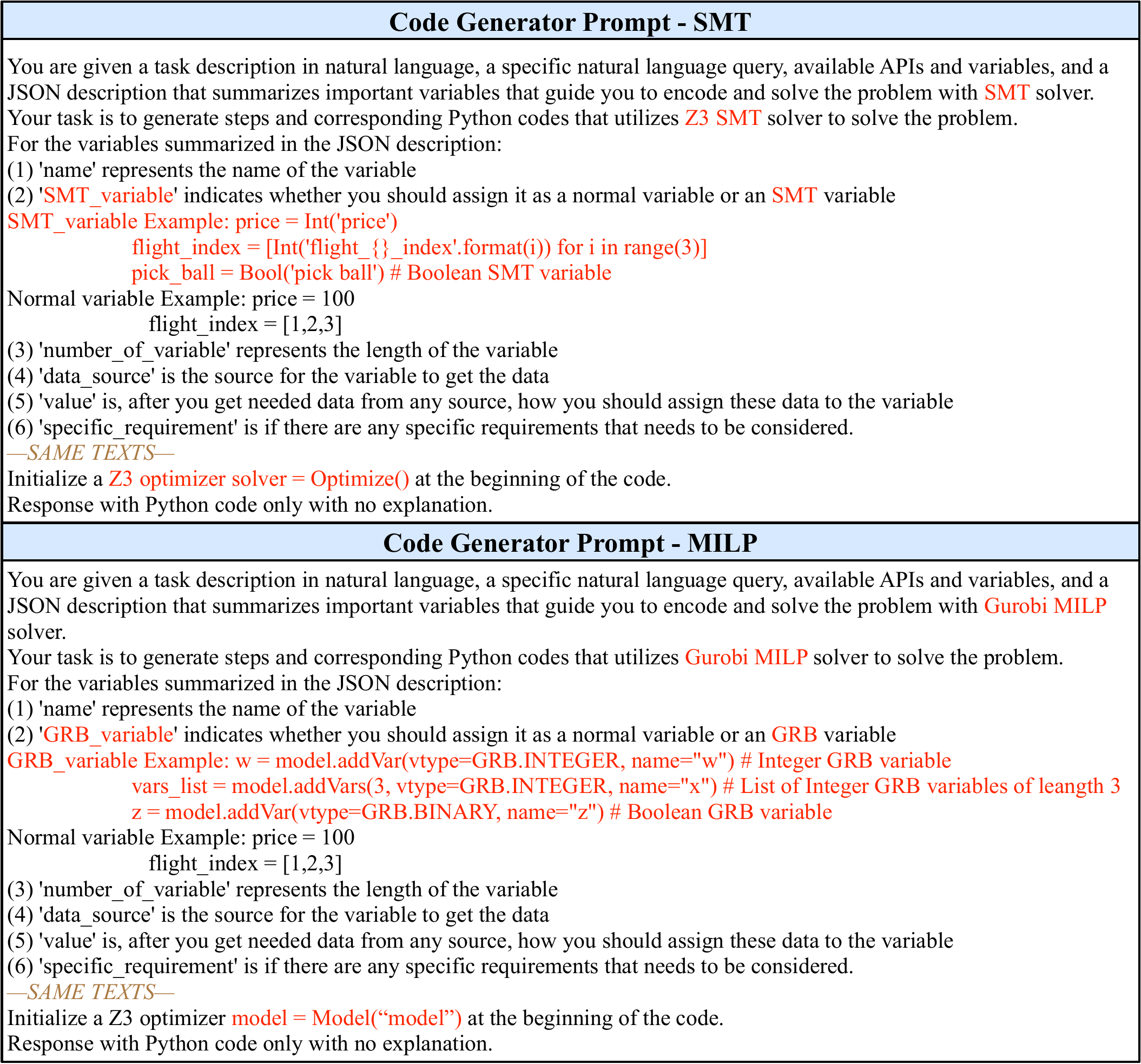} 
  \caption {The Code Genetator prompt difference when switching from using Z3 SMT solver to Gurobi MILP solver.}
  \label{fig:prompt_milp_2}
\end{figure}

\begin{figure}[!ht]
  \includegraphics[width=\linewidth]{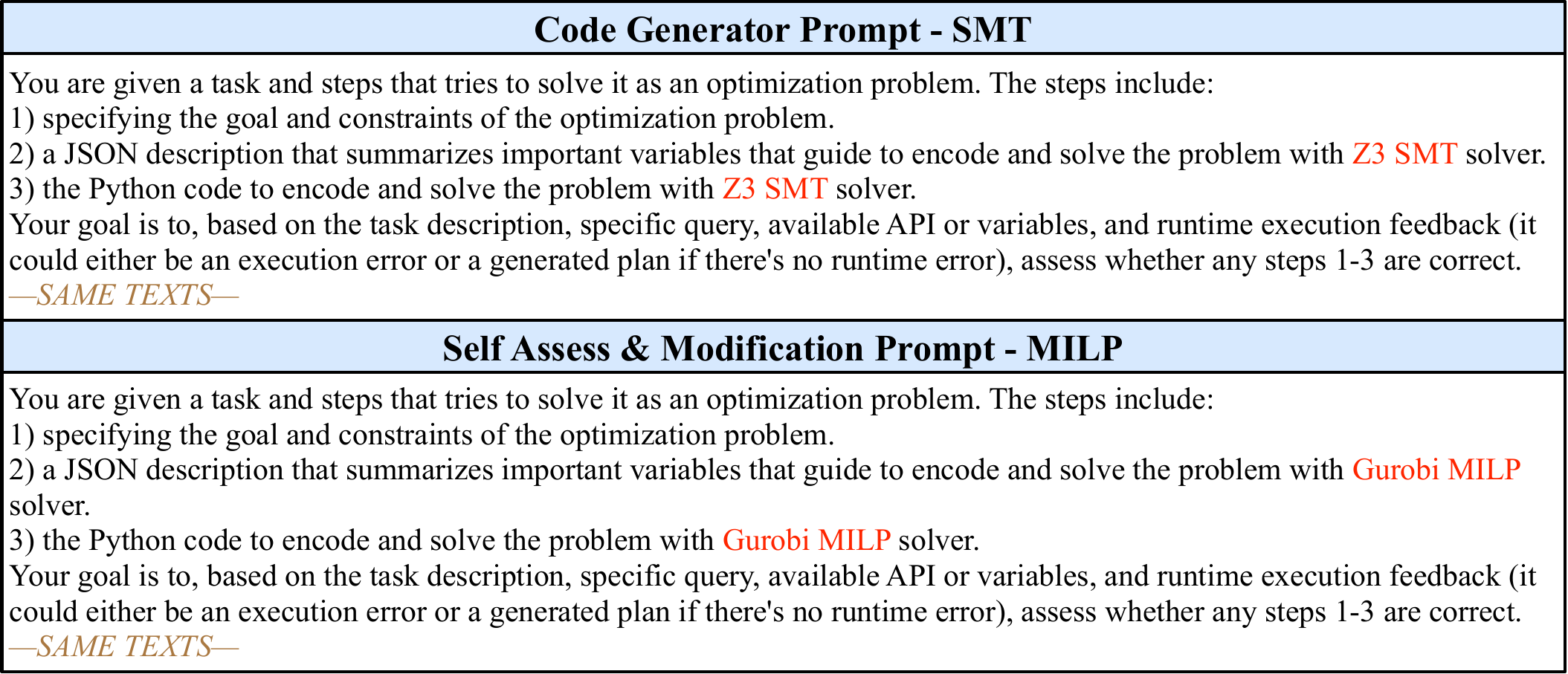} 
  \caption {The Self Assess \& Modification prompt difference when switching from using Z3 SMT solver to Gurobi MILP solver.}
  \label{fig:prompt_milp_3}
\end{figure}

\begin{figure}[!ht]
  \includegraphics[width=\linewidth]{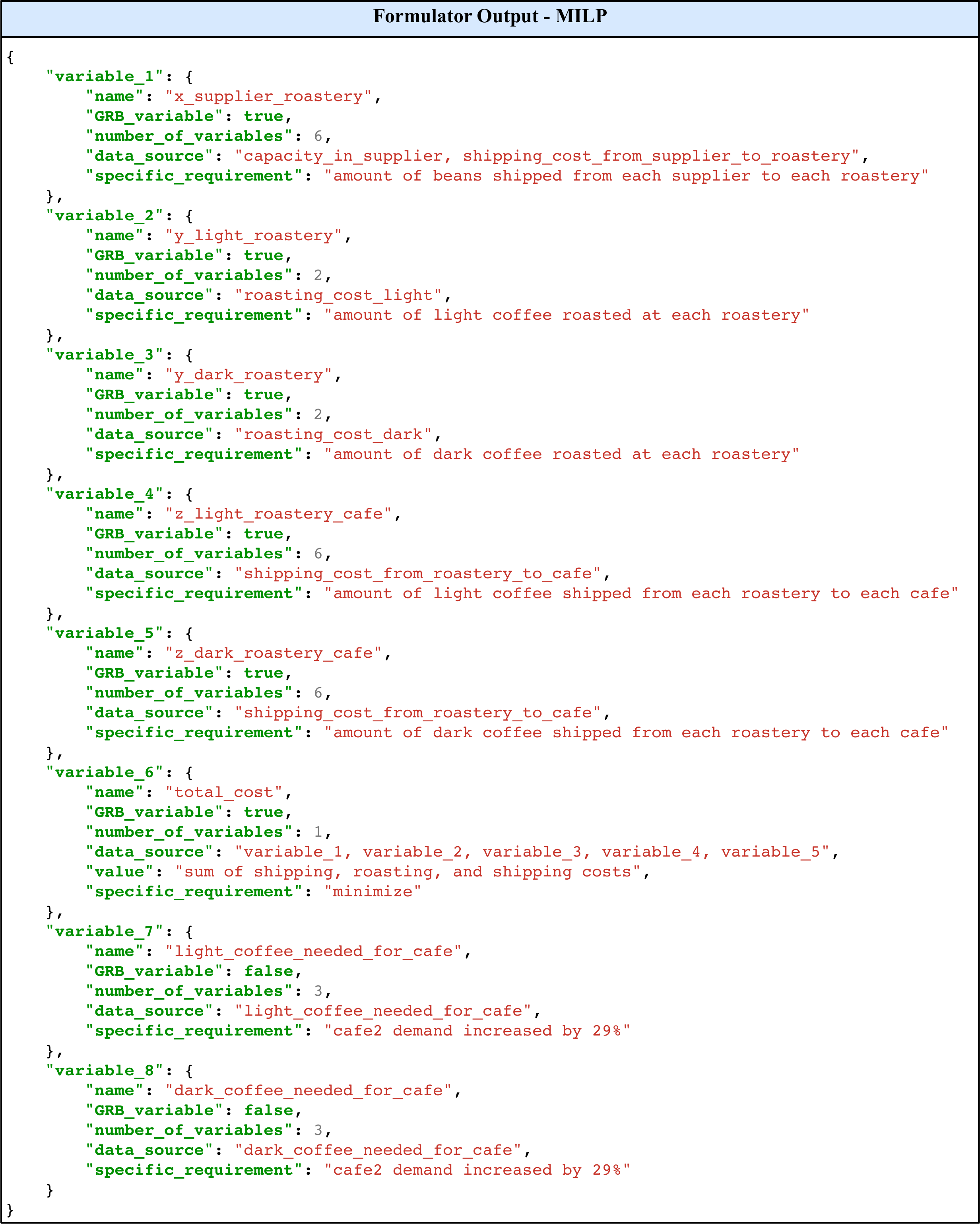}
  \caption {Output of Formulator for Coffee after switching the solver to MILP, using GPT-4o.}
  \label{fig:milp_output1}
\end{figure}

\begin{figure}[!ht]
  \includegraphics[width=\linewidth]{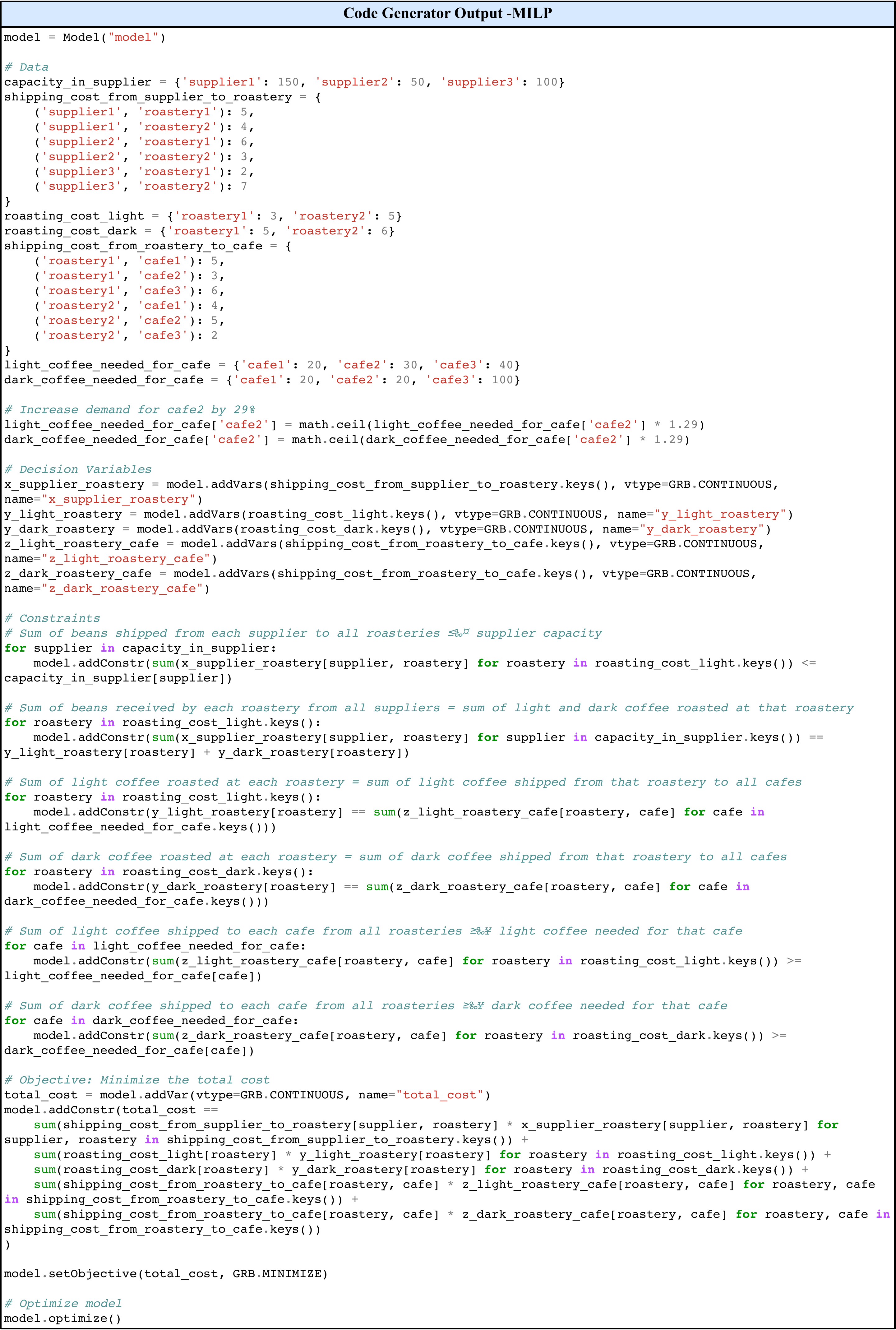}
  \caption {Output of Code Generator for Coffee after switching the solver to MILP, using GPT-4o.}
  \label{fig:milp_output2}
\end{figure}

We include the comparison of prompts for \textsc{Formulator}, \textsc{Code Generator}, and \textsc{Self Assess \& Modification} in Fig.~\ref{fig:prompt_milp_1} to~\ref{fig:prompt_milp_3} and labelled all the differences with red. We then include the output of \textsc{Formulator} and \textsc{Code Generator} in Fig.~\ref{fig:milp_output1} and Fig.~\ref{fig:milp_output2}.
The key takeaway is it is very easy to switch from one solver to another with \ours, as the inner logic is same: building an optimization problem.

\clearpage
\subsubsection{\blue{Paraphrased Prompts and Performance}}
\blue{To prove the flexibility of our prompt, we also paraphrase our NL task description and re-test the framework with a paraphrased description. The paraphrasing is performed by LLMs.}

\blue{One example paraphrased description is:}
\lstset{
    backgroundcolor=\color[RGB]{245,245,245},
    breaklines=true,
    basicstyle=\ttfamily\small,
    frame=trbl,
    frameround = tttt,
}\begin{lstlisting}
In this blocksworld problem, a robot arm can perform four actions: pickup, putdown, stack, and unstack. The environment consists of blocks that can be stacked, a single-block capacity arm, and a table.
Pickup: The arm can lift a block if it's clear, on the table, and the arm is empty. This results in the arm holding the block, which is no longer on the table or clear.
Putdown: If the arm is holding a block, it can place it on the table. This leaves the arm empty and the block on the table and clear.
Stack: The arm can place a block it's holding onto another clear block. This empties the arm, makes the top block clear and on the bottom block, while the bottom block becomes unclear.
Unstack: If a clear block is on another block and the arm is empty, it can lift the top block. This results in the arm holding the top block (no longer clear or on the bottom block), while the bottom block becomes clear.
\end{lstlisting}
\blue{With LLM-paraphrased random task descriptions, we test on 50 queries in Blockworld with Claude 3.5 Sonnet and shows LLMFP is still able to correctly generate 46/50 plans, reaching a high optimal rate of 92\%, significantly outperforming baselines. This shows our framework is not sensitive to the specific wordings of the task description, as long as they have adequate information. We can add more paraphrasing examples to show the robustness of LLMFP to different user inputs, if the reviewer finds it helpful to show the generalizability of LLMFP.}

\begin{table*}[!ht]
\caption{\blue{Optimal rate (\%) comparison of LLMFP with baselines with paraphrased prompts on Blocksworld with Claude 3.5 Sonnet}}
\label{paraphrase}
\begin{center}
\begin{small}
\begin{tabular}{ccccc}
\toprule
Direct$_{\textsc{GPT-4o}}$ &  CoT$_{\textsc{GPT-4o}}$ & Code$_{\textsc{GPT-4o}}$ & Code\_SMT$_{\textsc{GPT-4o}}$ & \ours$_{\textsc{GPT-4o}}$ \\
\midrule
32.0 & 46.0 & 0.0 & 0.0 & 92.0\\
\bottomrule
\end{tabular}
\end{small}
\end{center}
\end{table*}

\end{document}